\documentclass[letterpaper, 10 pt, conference]{ieeeconf}  % Comment this line out if you need a4paper

\IEEEoverridecommandlockouts                              % This command is only needed if 
\usepackage[colorlinks,linkcolor=blue]{hyperref} 
\usepackage{mathrsfs}
\usepackage{amsfonts}
\usepackage{amsmath}
\usepackage{multirow}
\usepackage{graphicx}
\usepackage{subfigure}
\usepackage{tikz}
\usetikzlibrary{calc,arrows,decorations.markings}
\usepackage{balance}
\usepackage{cite}
\usepackage{booktabs}
\usepackage{xpatch}
\usepackage{mathrsfs}
\usepackage{amsmath}
\usepackage{float}
\usepackage{placeins}
\usepackage{afterpage}
\usepackage{array}    % 用于调整列宽
\usepackage{tabularx}
\usepackage{titlesec}

\renewcommand{\arraystretch}{0.9}

\makeatletter
\patchcmd\@makecaption{\\}{.~}{}{\fail}
\makeatletter

\title{\LARGE \bf
VLAConf: Calibrated Task-Success Confidence for Vision-Language-Action Models}

\author{Dehao Huang$^{1,2}$, Aoxiang Gu$^{1}$, Chengjie Zhang$^{1}$, Bolin Zou$^{1}$,\\
Wenlong Dong$^{1}$, Zilang Cen$^{2,3}$, Yue Wang$^{2}$, and Hong Zhang$^{1}$\thanks{$^{1}$Department of Electronic and Electrical Engineering, Southern University of Science and Technology, Shenzhen, China.}\thanks{$^{2}$Zhongguancun Academy, Beijing, China.}\thanks{$^{3}$National Cybersecurity Academy, Wuhan University, Wuhan, China.}\thanks{This work is supported by the Zhongguancun Academy (Grant No. C20250203).}
}

\begin{document}

\maketitle

\begin{abstract}
Task-success confidence estimation for Vision-Language-Action (VLA) models provides a crucial task-level signal for monitoring manipulation in open-world environments and supporting downstream decision-making. Existing methods typically construct task-success confidence from action-token probabilities. However, such probabilities are not naturally available in flow-matching policies, limiting their applicability to mainstream flow-matching VLAs. To address this issue, we propose VLAConf, a two-stage representation-level confidence framework that operates on frozen pretrained VLA representations. A step-conditioned Coin-Flip Network learns an uncalibrated inverse success-support score from successful demonstrations, while a low-capacity calibrator fitted on outcome-labeled successful and failed rollouts maps the aggregated score to task-success probability. Experimental results on the LIBERO benchmark demonstrate that VLAConf improves online task-success confidence estimation over alternative approaches. We further demonstrate its utility in selective expert assistance, where confidence-triggered handoffs improve task success over no intervention. Its applicability is also evaluated in real-robot experiments.
\end{abstract}

\thispagestyle{empty}
\pagestyle{empty}

\section{INTRODUCTION}
Vision-Language-Action (VLA) models have demonstrated strong generalization across a wide range of robotic manipulation tasks by integrating pretrained vision-language representations with action generation~\cite{P012,P013,P014,P015}. However, deploying these policies in open-world environments remains challenging because changes in objects, scenes, and execution states can lead to unexpected failures. In addition to generating actions, a robot should therefore estimate how likely its ongoing execution is to accomplish the intended task. Task-success confidence provides such a task-level signal for monitoring manipulation and supporting downstream decision-making.

% Keep the motivation figure early so it floats to the first-page right column.
\begin{figure}[!t]
    \centering
    \includegraphics[width=0.92\linewidth,trim=1 1 1 1,clip]{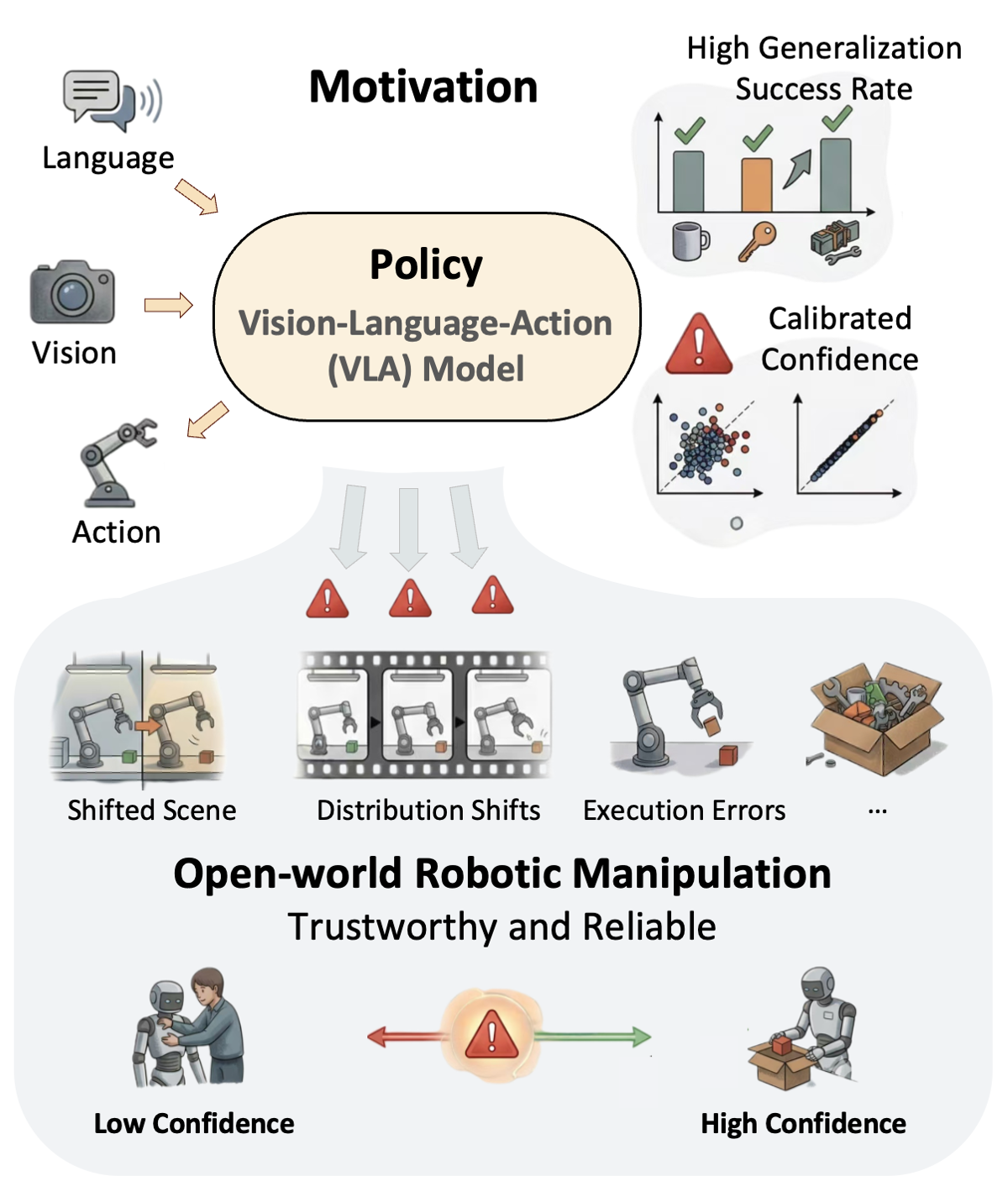}
    \vspace{-0.8em}
    \caption{Motivation of VLAConf. Despite strong generalization, VLA policies can still encounter shifted scenes, distribution shifts, and execution errors in open-world manipulation, highlighting the need for calibrated task-success confidence to distinguish low-confidence situations from high-confidence executions.}
    \vspace{-6mm}
    \label{fig:intro_overview}
\end{figure}

More specifically, task-success confidence estimates the probability that the current rollout will eventually succeed given the observed execution prefix and task instruction. As execution proceeds, this estimate is updated with newly observed states and task progress, allowing the robot to monitor whether the rollout remains likely to succeed. Once calibrated, it can support downstream decisions: the robot may continue autonomous execution when confidence remains high or request assistance when confidence becomes low~\cite{P001,P009,P029}.

Existing VLA confidence methods typically construct task-success estimates from action-token probabilities or agreement across sampled action outputs~\cite{zollo2025confidencevla}. This design is natural for autoregressive VLAs that expose normalized token distributions. However, mainstream flow-matching VLAs generate continuous action trajectories and do not naturally provide action-token probabilities~\cite{P012,P013}. Consequently, confidence estimators tied to discrete action outputs cannot be directly applied to these policies. Representation-based failure detection and one-class methods offer a possible alternative~\cite{P045,P048,gu2025safe}, but they have mainly been formulated as failure or out-of-distribution detectors rather than calibrated task-success estimators for generalist VLAs.

To address this issue, we propose VLAConf, a two-stage representation-level framework that estimates task-success confidence from frozen pretrained VLA representations. First, a step-conditioned Coin-Flip Network score head~\cite{lobel2023coinflip} learns an uncalibrated inverse success-support score from successful demonstrations. Step conditioning allows the same execution context to be evaluated relative to its phase in the manipulation process. The resulting scores are then aggregated over the observed rollout prefix, and a low-capacity calibrator fitted on outcome-labeled successful and failed rollouts maps the aggregated signal to task-success probability. In this way, VLAConf provides task-success confidence for flow-matching VLAs without relying on action-token probabilities.

Experimental results on LIBERO and its distribution-shift extensions, LIBERO-Plus and LIBERO-Pro, demonstrate that VLAConf improves online task-success confidence estimation over alternative approaches. On LIBERO-Goal, we further compare VLAConf with simple confidence predictors and representation-based methods, and evaluate whether the learned signal distinguishes successful and failed executions rather than only reflecting the average task success rate. To examine its downstream utility, we apply VLAConf to selective expert assistance on LIBERO-Goal, where confidence-triggered handoffs improve task success over no intervention. Its applicability is also evaluated in real-robot experiments.

In summary, the contributions of this work are three-fold:
\begin{itemize}
    \item We propose VLAConf, a two-stage representation-level framework that combines step-conditioned success-support scoring with outcome-supervised probability calibration and applies to flow-matching VLAs without requiring action-token probabilities.
    \item We evaluate VLAConf on simulated and real-world manipulation tasks, including comparisons with existing confidence estimators, simple confidence predictors, and representation-based methods.
    \item We demonstrate the downstream utility of VLAConf in a selective expert-assistance setting on LIBERO-Goal, where confidence-triggered handoffs improve task success over no intervention.
\end{itemize}

\section{Related Work} \label{related_work}

\subsection{Confidence Estimation and Uncertainty in VLAs}

Recent vision-language-action (VLA) models have demonstrated strong semantic grounding and broad manipulation capabilities by combining pretrained vision-language backbones with action generation modules \cite{P012,P013,P014,P015,P032}. Despite this progress, evaluating the reliability of VLA systems during deployment remains challenging. In the broader context of large language and vision-language models, uncertainty quantification often relies on token-level probabilities, semantic entropy, or generation consistency \cite{P017,P018,P019,P020,P022}. 

Existing VLA confidence and introspection methods commonly derive uncertainty signals from the policy's action outputs. ConfidenceVLA~\cite{zollo2025confidencevla}, for example, constructs task-success confidence from selected action-token probabilities and averages these estimates across paraphrased instructions. INSIGHT~\cite{karli2025insight} similarly uses sequences of token-level entropy and log-probability to predict when a VLA should request assistance. These approaches are naturally suited to autoregressive VLAs that expose normalized distributions over discrete action tokens. Mainstream flow-matching VLAs, however, generate continuous action trajectories and do not naturally provide comparable action-token probabilities~\cite{P012,P013,kwok2026cover}. Consequently, token-based confidence estimators cannot be directly applied to these policies. This limitation motivates estimating calibrated task-success confidence from the internal representations of flow-matching VLAs.

\subsection{Out-of-Distribution and Failure Detection in Robotics}

Extracting reliability signals from internal representations is deeply connected to out-of-distribution (OOD) and failure detection in robot manipulation. One practical regime limits the use of failure labels when learning a representation-level score because failed executions can be scarce, expensive, or unsafe to collect. Traditional OOD methods—such as density estimation, one-class discriminators, distance metrics, and consistency checks \cite{P045,P048,P057,P059,romer2025fiper,zhou2026rcnf,romer2025pathconsistent}—address this regime by modeling the support of successful executions without failure examples. Recent works, such as FAIL-Detect \cite{P048}, systematically evaluate these methods for generative imitation learning policies. VLAConf adopts successful-demonstration support modeling specifically for its representation-level score-learning stage, but prior methods have largely been designed for task-specific visuomotor policies. Scaling them to generalist VLAs introduces a significant multi-task generalization challenge, as the diverse visual scenes and novel language instructions inherent to generalist deployment can easily be misclassified as OOD anomalies.

To handle the complexity of generalist policies, a separate line of work explores supervised failure detection \cite{P050,P054,P058,P060}. For instance, SAFE \cite{gu2025safe} demonstrates that the internal latent features of VLAs capture rich, task-agnostic semantics about execution success, and trains a multi-task failure classifier using outcome-labeled successful and failed rollouts. This provides direct supervision but requires diverse failed executions to train the representation-level detector. VLAConf separates the two roles: successful demonstrations train the representation-level support scorer, whereas outcome-labeled successful and failed rollouts fit a low-capacity probability calibrator rather than the representation-level scorer.

\textbf{Positioning of Our Approach.} VLAConf combines successful-demonstration support modeling with outcome-supervised probability calibration to provide task-success confidence for flow-matching VLAs. A lightweight Coin-Flip Network head learns an uncalibrated inverse success-support score from successful demonstrations, while a low-capacity calibrator maps the aggregated score to task-success probability using a small set of outcome-labeled successful and failed rollouts. Because the score is constructed from frozen internal representations rather than action-token distributions, VLAConf can be applied to flow-matching policies that do not expose action-token probabilities.

\section{Problem Definition}\label{problem}

\begin{figure}[!t]
    \centering
    \includegraphics[width=\linewidth]{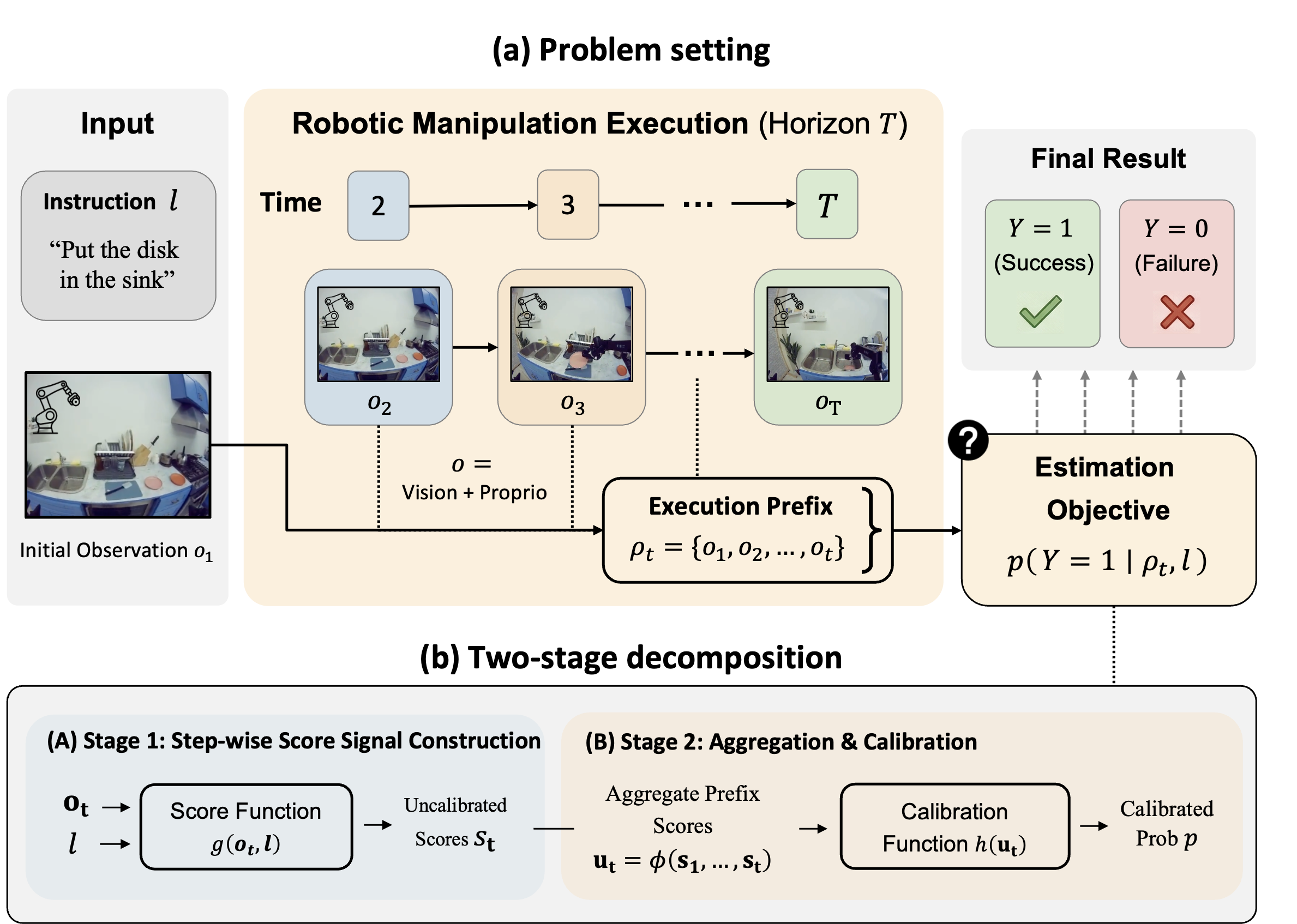}
    \vspace{-6mm}
    \caption{Problem formulation for VLA confidence estimation. Top: given an instruction and an observed rollout prefix, the goal is to estimate the probability that the current execution will eventually succeed. Bottom: VLAConf decomposes this problem into step-wise score construction followed by prefix aggregation and calibration.}
    \vspace{-6mm}
    \label{fig:problem_formulation}
\end{figure}

We study task-success confidence estimation for deployed vision-language-action (VLA) policies in language-conditioned robotic manipulation, as shown in the upper part of Figure~\ref{fig:problem_formulation}. Given a language instruction $l$ and an observed execution prefix $\rho_t=(o_1,\dots,o_t)$, where $o_t=(I_t,x_t)$ contains the current image and proprioceptive state, the goal is to estimate the probability that the current rollout will eventually succeed under the deployed policy, $p(Y=1\mid\rho_t,l)$, where $Y\in\{0,1\}$ denotes the final rollout-level task outcome. This formulation covers both \textsc{pre-execution} confidence, where only the initial observation is available, and \textsc{online execution} confidence, where the estimate is updated as the rollout unfolds. The latter is important for manipulation because task progress and emerging failures are often only visible after execution begins.

We assume a practical imitation-learning workflow. The VLA policy is first pretrained on broad robot data and then adapted to the target task using successful demonstrations, denoted by $\mathcal{D}_{\mathrm{sft}}$. To learn the representation-level inverse success-support score, we use the observation-language traces of these demonstrations, written as $\tau_i=(o_{i,1:T_i}, l_i)$, where $l_i$ is the language instruction and $o_{i,k}=(I_{i,k},x_{i,k})$. Action labels used for policy adaptation are omitted from the notation because VLAConf keeps the policy parameters fixed and learns the score on top of frozen VLA representations. These trajectories originate from policy adaptation rather than a separate confidence-data collection and contain no failed executions.

In addition, a small set of completed policy rollouts, containing both successful and failed executions, can be labeled with binary task outcomes; we denote this outcome-labeled set by $\mathcal{D}_{\mathrm{cal}}=\{(\rho^{(j)},l^{(j)},Y^{(j)})\}_{j=1}^{N_{\mathrm{cal}}}$. The two datasets play complementary roles: $\mathcal{D}_{\mathrm{sft}}$ defines the support of phase-conditioned successful execution and is used to learn an uncalibrated inverse support signal, while $\mathcal{D}_{\mathrm{cal}}$ identifies how the aggregated signal maps to task-success probability.

As summarized in the lower part of Figure~\ref{fig:problem_formulation}, VLAConf first produces step-wise inverse success-support scores and aggregates the observed prefix into a scalar signal $u_t$. It then fits a calibration map $h(\cdot)$, implemented as Platt scaling in our experiments, to convert this signal into $p(Y=1\mid\rho_t,l)=h(u_t)$.

\section{VLAConf}\label{method}

\subsection{Overview}

An overview of VLAConf is shown in Figure~\ref{fig:vlaconf_overview}. Given a rollout prefix $\rho_t=(o_1,\dots,o_t)$ and instruction $l$, VLAConf estimates task-success confidence by measuring whether the observed execution remains close to the success manifold represented by $\mathcal{D}_{\mathrm{sft}}$. The method first extracts a step-wise execution representation from a frozen VLA, maps each step to a phase-conditioned \emph{inverse success-support score}---an inverse-pseudocount signal that measures how strongly the current execution is covered by successful demonstrations---and finally aggregates and calibrates these scores into $p(Y=1\mid\rho_t,l)$.

We first describe how VLAConf constructs backbone-agnostic execution features from frozen VLA hidden states. We then introduce the step-conditioned success-support scoring objective trained on successful trajectories. Finally, we present prefix-level aggregation and post-hoc calibration, which convert step-wise inverse success-support scores into calibrated task-success probabilities.

\subsection{Frozen VLA Execution Representation}

The first step is to obtain a representation on which support can be estimated consistently across VLA architectures. We estimate confidence in the \emph{internal representation space} of a pretrained VLA rather than in its action-output space. This is important because manipulation failures are frequently reflected by whether the current multimodal context departs from familiar successful behavior, not merely by whether the immediate action prediction appears sharp. Moreover, action probabilities are often unavailable or hard to compare across architectures, especially when the action head is continuous rather than token based.

Let $f_{\theta}$ denote a pretrained VLA backbone. We keep $f_{\theta}$ frozen throughout confidence learning. This choice preserves the semantic and geometric priors already learned by the VLA and keeps confidence estimation efficient because only a small head is optimized. Thus, VLAConf treats the pretrained VLA as a general-purpose multimodal feature extractor and learns confidence on top of that representation rather than by adapting the policy itself.

At step $k$, the backbone processes the current observation-instruction pair $(o_k,l)$ and returns hidden states $\mathcal{H}_k$. Although different backbones expose these hidden states through different tensor layouts, VLAConf only assumes a shared pooled-feature interface. We separately pool the valid visual tokens and language tokens, yielding a visual summary $h_k^v$ and a language summary $h_k^l$. These summaries capture the scene content and task semantics needed for confidence estimation. We additionally fuse the robot proprioceptive state because visual-language context alone may not fully determine whether the current execution phase is healthy. Variables such as gripper opening and end-effector pose often distinguish between a normal state and an abnormal one even when the image appears similar. We project the proprioceptive component $x_k$ with a small state encoder and fuse it with the pooled visual-language features:
\[
\begin{aligned}
\mathcal{H}_k &= f_{\theta}(o_k, l), \\
h_k^v &= \mathrm{Pool}_v(\mathcal{H}_k), \qquad
h_k^l = \mathrm{Pool}_l(\mathcal{H}_k), \\
h_k^x &= E_x(x_k), \qquad
z_k = M_{\text{mix}}([h_k^v; h_k^l; h_k^x]),
\end{aligned}
\]
where $\mathrm{Pool}_v(\cdot)$ and $\mathrm{Pool}_l(\cdot)$ denote masked mean pooling over the valid visual and language token positions, $E_x(\cdot)$ is a small MLP state encoder, and $M_{\text{mix}}(\cdot)$ is a two-layer mixing MLP. The result is a compact execution representation $z_k$ on which confidence is learned. This representation hides backbone-specific implementation details while retaining the visual, semantic, and proprioceptive information needed to assess familiarity.

\subsection{Step-Conditioned Success-Support Scoring}

Given the execution feature $z_k$, we seek a scalar score that measures how strongly the current execution is supported by successful behavior. We construct this score using a Coin-Flip Network (CFN)~\cite{lobel2023coinflip}, which turns random-label regression into a representation-space pseudocount. Because the meaning of an execution state depends on when it occurs, we further condition the score on the current rollout step. For example, an open gripper is expected before grasping but may indicate an unsuccessful execution after lifting. The resulting score therefore measures support relative to the corresponding manipulation phase.

To encode task phase, we introduce a step encoder that maps the rollout step index $k$ to a compact descriptor. Let $K$ be a fixed normalization horizon and let $\hat{k}=\mathrm{clip}(k{-}1,0,K{-}1)$. The descriptor combines a learned step embedding with a normalized progress scalar:
\[
\psi_k = [\,e_k;\, \bar{k}\,], \qquad e_k = \mathrm{Emb}(\hat{k}), \qquad \bar{k} = \hat{k}/(K{-}1),
\]
where $\mathrm{Emb}(\cdot)$ is a learned embedding table of size $K$. Using a fixed $K$ rather than the per-trajectory length $T_i$ keeps $\bar{k}$ well-defined during online execution, where the eventual rollout length is unknown. The clipping operation keeps both $e_k$ and $\bar{k}$ in range when $k$ exceeds $K{-}1$.

% Queue the method overview near the method section.
\afterpage{%
\begin{figure}[!t]
    \centering
    \includegraphics[width=\columnwidth,trim=15 15 15 0,clip]{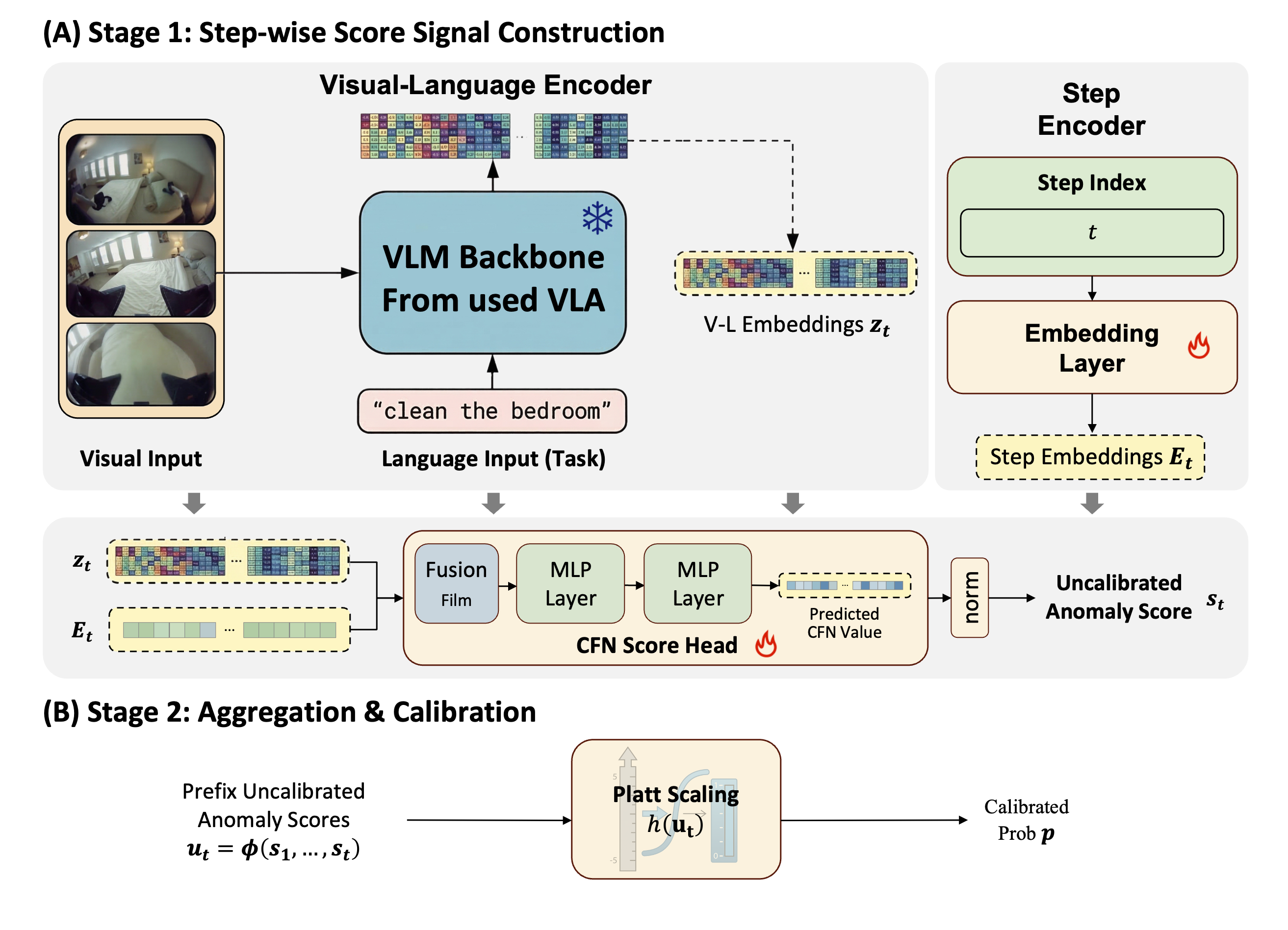}
    \vspace{-5mm}
    \caption{Overview of VLAConf. For each observed step, Stage 1 extracts frozen VLA features and applies a step-conditioned head to construct a step-wise inverse success-support score $s_k = g(o_k,l,k)$. Over a rollout prefix, Stage 2 aggregates the available scores into an uncalibrated prefix-level signal $u_t = \phi(s_1,\dots,s_t)$ and converts it into task-success confidence by post-hoc calibration, $p(Y=1 \mid \rho_t,l)=h(u_t)$. The full model's proprioceptive fusion branch is omitted for clarity.}
    \label{fig:vlaconf_overview}
    \vspace{-5mm}
\end{figure}
}

Instead of concatenating $\psi_k$ with $z_k$, we use the descriptor to modulate the scoring head through a FiLM-style affine transform~\cite{perez2018film} followed by a residual gate. This allows the same execution representation to be evaluated against different phase-specific successful behaviors. Let $W_{\mathrm{proj}}(\cdot)$ denote a $\mathrm{Linear}{\to}\mathrm{LN}{\to}\mathrm{Tanh}$ pre-projection, $A_{\mathrm{cond}}(\cdot)$ a two-layer MLP that encodes $\psi_k$, and $q_{\phi}(\cdot)$ a two-layer coin-value head. We write
\[
\begin{aligned}
h_k &= W_{\mathrm{proj}}(z_k), \qquad \chi_k = A_{\mathrm{cond}}(\psi_k), \\
(\gamma_k,\beta_k) &= W_{\mathrm{film}}(\chi_k), \qquad \eta_k = \sigma\!\big(W_{\mathrm{gate}}(\chi_k)\big), \\
\tilde{h}_k &= \mathrm{LN}\!\big((1+\gamma_k)\odot h_k + \beta_k\big), \\
\tilde{z}_k &= \eta_k\odot\tilde{h}_k + (1-\eta_k)\odot h_k, \\
v_k &= q_{\phi}(\tilde{z}_k), \qquad s_k = \|v_k\|_2^2,
\end{aligned}
\]
where $\sigma(\cdot)$ is the sigmoid function, $\odot$ denotes element-wise multiplication, and $W_{\mathrm{film}}$ and $W_{\mathrm{gate}}$ are independent linear maps that share the upstream conditioning network $A_{\mathrm{cond}}$. The post-affine LayerNorm stabilizes the modulated branch before gating, and the residual gate balances the phase-conditioned feature $\tilde h_k$ with the original projected feature $h_k$.

We train the scoring head on step-wise samples extracted exclusively from successful demonstrations:
\[
\mathcal{D}_{\mathrm{step}}^{+}
=\{(o_{i,k},l_i,k):\tau_i\in\mathcal{D}_{\mathrm{sft}},\; k=1,\dots,T_i\}.
\]
Each sample is assigned a fixed pseudo-random Rademacher vector $c_{i,k}=\Gamma(i,k)\in\{-1,+1\}^{d}$, whose entries are generated independently across sample identifiers. The scoring head is optimized by
\[
\mathcal{L}_{\mathrm{CFN}}
=\frac{1}{|\mathcal{D}_{\mathrm{step}}^{+}|}
\sum_{\tau_i\in\mathcal{D}_{\mathrm{sft}}}\sum_{k=1}^{T_i}
\|q_{\phi}(\tilde z_{i,k})-c_{i,k}\|_2^2.
\]

The CFN objective provides an inverse-count interpretation of the learned score. Consider a representation pattern $\xi$ with $n_{\xi}$ occurrences and independently generated targets $c_1,\ldots,c_{n_{\xi}}$. At the squared-error optimum, the prediction for $\xi$ is the empirical mean of its targets:
\[
q_{\phi}^{*}(\xi)=\frac{1}{n_{\xi}}\sum_{r=1}^{n_{\xi}}c_r.
\]
Because every target dimension is an independent zero-mean, unit-variance Rademacher variable,
\[
\mathbb{E}_{c}\!\left[\frac{1}{d}\left\|q_{\phi}^{*}(\xi)\right\|_2^2\right]
=\frac{1}{n_{\xi}}.
\]
Thus, random-label regression produces a score whose expected magnitude is inversely proportional to representation count. Through parameter sharing across related execution features, the learned CFN induces an effective pseudocount $\widehat N_{+}(\tilde z_k)$ in the representation space of successful demonstrations. We therefore use
\[
s_k=\|q_{\phi}(\tilde z_k)\|_2^2
\;\propto\;
\frac{1}{\widehat N_{+}(\tilde z_k)}
\]
as the step-wise inverse success-support score. The factor $1/d$ is omitted because $d$ is fixed and only changes the global scale of the score.

The inverse pseudocount becomes relevant to task success under a phase-conditioned success-support premise. For a fixed instruction $l$ and execution phase $k$, let $m_{+}(\xi\mid l,k)$ denote the local probability mass around representation $\xi$ under the distribution of successful executions. Consider two rollout prefixes $\rho_k^{a}$ and $\rho_k^{b}$ at the same task phase, with corresponding representations $\tilde z_k^{a}$ and $\tilde z_k^{b}$. We assume that stronger support under the successful-execution distribution indicates a higher probability of successful continuation:
\[
\begin{aligned}
&m_{+}(\tilde z_k^{a}\mid l,k)
\ge m_{+}(\tilde z_k^{b}\mid l,k) \\
&\qquad\Longrightarrow
P_{\pi}(Y=1\mid\rho_k^{a},l)
\ge P_{\pi}(Y=1\mid\rho_k^{b},l).
\end{aligned}
\]

This premise follows naturally from task-adapted imitation learning. The policy is trained to reproduce the successful trajectories in $\mathcal{D}_{\mathrm{sft}}$, so high-support regions correspond to execution configurations from which successful behavior has repeatedly been demonstrated. For example, during the grasp phase, demonstrations consistently place the end effector near the target while the gripper transitions toward closure. An execution with the same spatial and proprioceptive configuration is therefore well supported, whereas a closed gripper far from the target is weakly supported and provides less evidence of a successful continuation. Because the VLA representation combines the observation, instruction, and proprioceptive state, and the step-conditioned transformation compares these features at the corresponding manipulation phase, its local support captures this task- and phase-specific notion of execution consistency.

For a fixed successful dataset, the local empirical count is proportional to $m_{+}$, while CFN estimates its inverse through $s_k$. The successful-support premise therefore induces the ordering
\[
s_k^{a}\le s_k^{b}
\quad\Longrightarrow\quad
P_{\pi}(Y=1\mid\rho_k^{a},l)
\ge
P_{\pi}(Y=1\mid\rho_k^{b},l).
\]
Thus, the CFN score provides a local, phase-conditioned signal of successful continuation. Since rollout success depends on the sequence of observed execution states rather than a single step, we next aggregate these local scores over the rollout prefix.

\subsection{Prefix Aggregation and Calibration}

We aggregate the local inverse-support scores over the observed prefix:
\[
u_t=\phi(s_1,\dots,s_t),
\]
where $\phi$ is coordinate-wise non-decreasing, so increasing any constituent inverse-support score cannot reduce the aggregated signal. In the pre-execution setting, only the initial observation is available and we set $u_1=s_1$. During online execution, we consider the running mean, $u_t=\tfrac{1}{t}\sum_{k=1}^{t}s_k$, and the maximum prefix score. Both summarize the accumulated success-support evidence while preserving its ordering.

The aggregated signal $u_t$ is converted into an interpretable success probability using a small set of outcome-labeled successful and failed policy rollouts, $\mathcal{D}_{\mathrm{cal}}$, introduced in Section~\ref{problem}. This is the only estimator-training stage that consumes rollout outcomes, and the CFN scoring head remains fixed. We fit a Platt-scaling model~\cite{P005}:
\[
h(u_t)=\operatorname{sigmoid}(-\alpha u_t+\beta),
\]
where $\alpha\ge 0$ and $\beta\in\mathbb{R}$ are learned by minimizing binary cross-entropy on $\mathcal{D}_{\mathrm{cal}}$. The non-negative slope preserves the successful-support ordering: a larger aggregated inverse-support score is mapped to a lower probability of task success. Overall, CFN provides the representation-level inverse success-support signal, step conditioning makes this signal phase-aware, and prefix calibration converts it into rollout-level task-success confidence.

\begin{table*}[!t]
    \centering
    \caption{Main comparison averaged over standard LIBERO suites.}
    \vspace{-1mm}
    \label{tab:confvla_alignment}
    \footnotesize
    \setlength{\tabcolsep}{2.5pt}
    \renewcommand{\arraystretch}{0.98}
    \begin{tabular*}{\textwidth}{@{\extracolsep{\fill}}llcccccccc@{}}
    \toprule
    \multirow{2}{*}{\textbf{Backbone}} &
    \multirow{2}{*}{\textbf{Method}} &
    \multicolumn{3}{c}{\textbf{Pre-Execution}} &
    \multicolumn{3}{c}{\textbf{Online Execution}} &
    \multirow{2}{*}{\textbf{Succ. (\%)}} &
    \multirow{2}{*}{\textbf{Avg. Inf. Time (ms)}} \\
    \cmidrule(lr){3-5} \cmidrule(lr){6-8}
    & & \textbf{ECE $\downarrow$} & \textbf{Brier $\downarrow$} & \textbf{NLL $\downarrow$} & \textbf{ECE $\downarrow$} & \textbf{Brier $\downarrow$} & \textbf{NLL $\downarrow$} & & \\
    \midrule
    \multirow{5}{*}{OpenVLA-OFT}
    & PCA-kmeans       & 0.3031 & 0.3103 & 1.0976 & 0.3689 & 0.3510 & 1.1169 & 76.9 & 63.7 \\
    & TokenProb        & 0.2123 & 0.2220 & 0.6352 & 0.0922 & 0.1764 & 0.5404 & 77.7 & 65.1 \\
    & ConfidenceVLA    & 0.0363 & 0.1702 & 0.5295 & \textbf{0.0276} & 0.1647 & 0.5041 & 77.7 & 712.9 \\
    & VLAConf-NoStep   & 0.0410 & 0.1691 & 0.5148 & 0.0661 & 0.1656 & 0.5063 & 76.9 & 64.5 \\
    & VLAConf          & \textbf{0.0340} & \textbf{0.1614} & \textbf{0.4991} & 0.1188 & \textbf{0.1073} & \textbf{0.3335} & 78.2 & 64.9 \\
    \midrule
    \multirow{4}{*}{\boldmath$\pi^{0.5}$}
    & PCA-kmeans       & 0.4168 & 0.3428 & 1.1049 & 0.4053 & 0.3394 & 1.0630 & 91.3 & 163.5 \\
    & ConfidenceVLA    & / & / & / & / & / & / & / & / \\
    & VLAConf-NoStep   & 0.0579 & 0.0898 & 0.3387 & \textbf{0.0448} & 0.0884 & 0.3261 & 90.5 & 164.4 \\
    & VLAConf          & \textbf{0.0370} & \textbf{0.0821} & \textbf{0.3141} & 0.0515 & \textbf{0.0668} & \textbf{0.2501} & 91.3 & 164.7 \\
    \bottomrule
    \end{tabular*}
    \vspace{-4mm}
\end{table*}
% /home/hdh/projects/ConfidenceRLinf/logs/0_base_exp/logs/batch_eval_with_critic_20260325-03:07:20/final_metrics_suite-final_metrics.log
% /home/hdh/projects/ConfidenceRLinf/logs/0_base_exp/logs/batch_eval_openvla_oft_with_critic_20260324-13:53:01/final_metrics_suite-final_metrics.log
% /home/hdh/projects/ConfidenceRLinf/logs/0_base_exp/logs/calibration/20260326-02:14:35_baseline_batch

\section{Experiments} \label{exp}

\subsection{Experimental Setup}\label{exp:setup}

\noindent\textbf{Evaluation organization.} Our evaluation examines three aspects of VLAConf. First, we evaluate task-success confidence before and during manipulation, including its robustness under distribution shifts. Second, we study whether the estimated confidence can support selective expert assistance while balancing task success and expert usage. Third, we evaluate its applicability to real-world manipulation.

\noindent\textbf{Backbones.} We evaluate VLAConf on two representative VLA backbones spanning the discrete- and continuous-action regimes: (i) \textsc{OpenVLA-OFT}~\cite{P012}, a token-based VLA that produces actions through a discrete autoregressive head; and (ii) \textsc{$\pi^{0.5}$}~\cite{P013}, a state-of-the-art generalist VLA that emits continuous actions via flow-matching. Both backbones are kept frozen during confidence learning. This pair lets us examine whether the same confidence signal can be constructed across heterogeneous action heads, and is also the setting under which prior action-token-based confidence methods break down.

\noindent\textbf{Benchmarks.} We use the LIBERO benchmark suite~\cite{liu2023libero} for in-distribution evaluation, covering three subsets that stress different aspects of manipulation: \textsc{LIBERO-Goal}, \textsc{LIBERO-Spatial}, and \textsc{LIBERO-Object}. For robustness under shifted conditions, we additionally evaluate on \textsc{LIBERO-Pro}\cite{zhou2025libero_pro} and \textsc{LIBERO-Plus}\cite{fei2025libero_plus}, which extend LIBERO with perturbations and previously unseen tasks (Section~\ref{exp:generalization}). Real-robot results are reported in Section~\ref{exp:realworld}.

\noindent\textbf{Baselines.} We compare VLAConf with reference methods spanning action outputs, simple predictors, and frozen representations. (i) \textsc{TokenProb} aggregates the policy's own action-token probabilities over the action sequence. (ii) \textsc{ConfidenceVLA}~\cite{zollo2025confidencevla} estimates task-success probability from action-token agreement across resampled prompts. Both methods require per-token action probabilities and are therefore unavailable for continuous-action backbones such as $\pi^{0.5}$; the corresponding entries are marked ``/'' in Table~\ref{tab:confvla_alignment}. (iii) \textsc{Success Prior} predicts the calibration-set success frequency for every rollout. (iv) \textsc{Logistic Regression} is a supervised linear predictor fit to frozen VLA representations using the outcome-labeled calibration split. (v) \textsc{PCA-kmeans}, adapted from robot execution failure detection~\cite{liu2024fleetworldmodels}, projects frozen VLA features with PCA and uses the distance to the nearest successful-feature cluster as an anomaly score. (vi) \textsc{DeepSVDD}~\cite{ruff2018deeponeclass} is a one-class representation baseline trained on successful demonstration features. We additionally evaluate \textsc{VLAConf-NoStep}, which retains the same frozen features and success-support score-learning objective as VLAConf but removes the step encoder, corresponding to an unconditioned support scorer trained on successful demonstrations~\cite{burda2018exploration,yang2024consistencyflowmatching}.

\noindent\textbf{Evaluation protocol.} We split completed rollouts into an outcome-labeled calibration set $\mathcal{D}_{\mathrm{cal}}$, containing both successful and failed executions, and a disjoint test set. On $\mathcal{D}_{\mathrm{cal}}$, VLAConf fits only two Platt-scaling parameters, whereas Logistic Regression additionally learns its linear prediction weights. We evaluate confidence in two settings: \textsc{pre-execution}, using only the initial observation $o_1$, and \textsc{online execution}, using the observed rollout prefix $\rho_t$.

\noindent\textbf{Metrics.} We report three calibration metrics that are standard for binary outcomes and capture complementary aspects of probability quality: \textsc{Expected Calibration Error} (ECE) measures the average gap between predicted confidence and empirical accuracy across confidence bins; \textsc{Brier score} measures the mean squared error between predicted probability and the binary outcome; and \textsc{negative log-likelihood} (NLL) penalizes overconfident wrong predictions more aggressively than Brier. Lower is better for all three. Because calibration error alone can favor a nearly constant predictor when policy success is high, the $\pi^{0.5}$ comparison also reports \textsc{Failure AUROC}, treating failure as the positive class, to measure how well confidence ranks failed versus successful rollouts. We additionally report task \textsc{Success Rate} of the underlying policy and the \textsc{Average Inference Time} per confidence query, measured on the same hardware.

\begin{figure}[!t]
    \centering
    \includegraphics[width=0.92\linewidth]{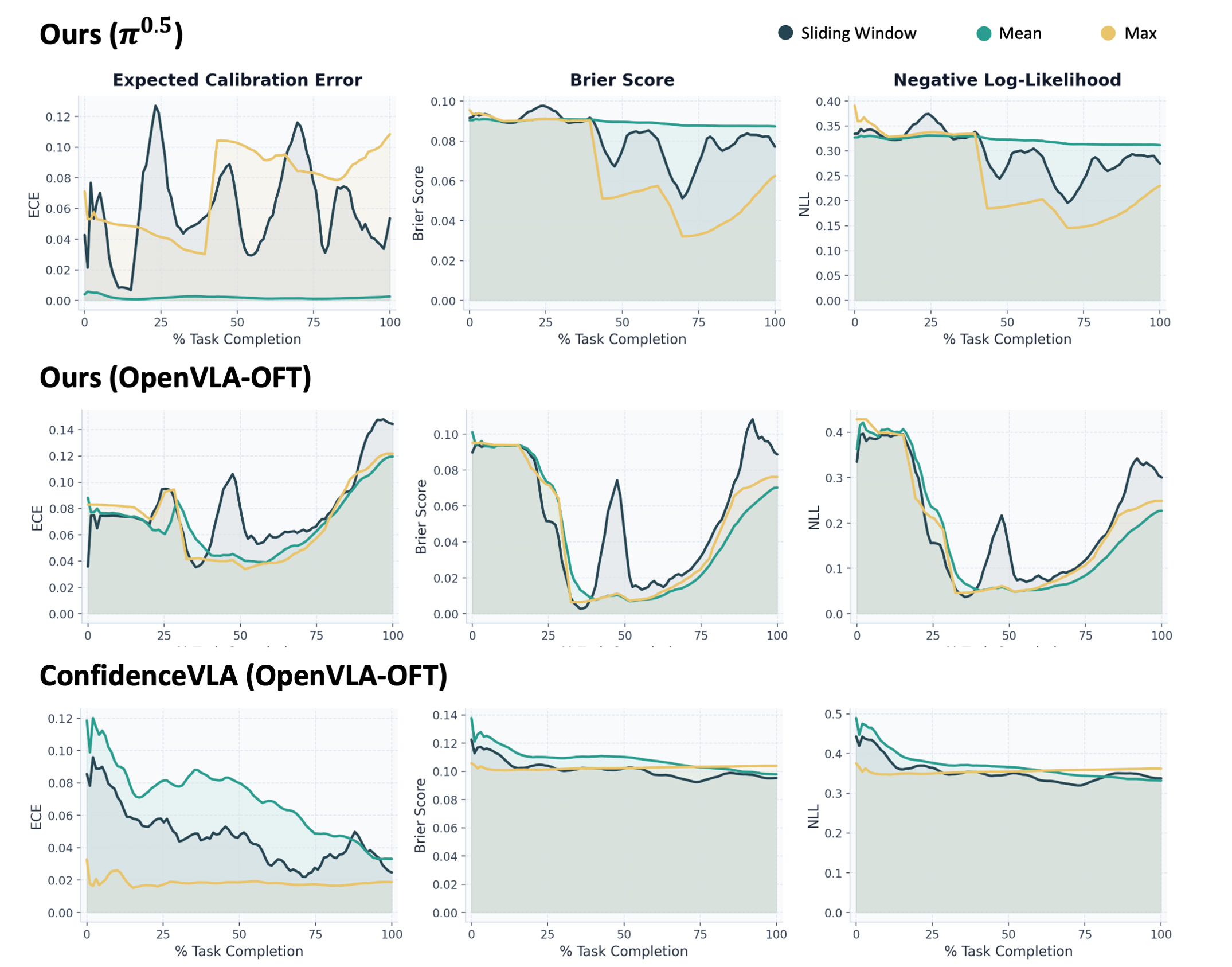}
    \caption{Online calibration trends across task completion. ECE, Brier score, and NLL are evaluated across prefix lengths and aggregation rules.}
    \label{fig:online_temporal_calibration}
    \vspace{-5mm}
\end{figure}
% /home/hdh/projects/ConfidenceRLinf/logs/0_base_exp/logs/batch_eval_openvla_oft_with_critic_20260324-13:53:01/goal_step2000/final_metrics_suite-final_metrics/success_once_across_time_platt/across_time_dashboard.png
% /home/hdh/projects/ConfidenceRLinf/logs/0_base_exp/logs/calibration/20260326-02:14:35_baseline_batch/goal/baseline/baseline_final_metrics_suite-final_metrics/success_once_across_time_platt/across_time_dashboard.png
% /home/hdh/projects/ConfidenceRLinf/logs/0_base_exp/logs/batch_eval_with_critic_20260325-03:07:20/goal_step2000/final_metrics_suite-final_metrics/success_once_across_time_platt/across_time_dashboard.png

\subsection{Main Results}\label{exp:main}

We first evaluate on the standard LIBERO suites, which provide a controlled in-distribution setting for comparing confidence signals across backbones and execution regimes. Table~\ref{tab:confvla_alignment} reports the average comparison over \textsc{LIBERO-Spatial}, \textsc{LIBERO-Object}, and \textsc{LIBERO-Goal}; the full suite-level results are provided in Appendix Table~\ref{tab:libero_standard_full}. To examine whether probability quality is accompanied by useful failure ranking, Table~\ref{tab:additional_confidence_baselines} compares VLAConf with simple, supervised, and representation-based references on $\pi^{0.5}$. Within each comparison, bold marks the best applicable metric. Small success-rate differences reflect stochasticity from independently repeated rollouts; the confidence estimators do not modify the underlying policy. Average inference time is measured per confidence query on the same hardware. We highlight four observations.

\begin{table}[!t]
    \centering
    \caption{Online confidence comparison averaged over standard LIBERO suites.}
    \label{tab:additional_confidence_baselines}
    \footnotesize
    \setlength{\tabcolsep}{2.2pt}
    \renewcommand{\arraystretch}{1.03}
    \begin{tabular*}{\columnwidth}{@{\extracolsep{\fill}}lcccc@{}}
    \toprule
    \textbf{Method} & \textbf{ECE $\downarrow$} & \textbf{Brier $\downarrow$} & \textbf{NLL $\downarrow$} & \textbf{Fail. AUROC $\uparrow$} \\
    \midrule
    Success Prior       & \textbf{0.017} & 0.094 & 0.336 & 0.500 \\
    Logistic Regression & 0.056 & 0.072 & \textbf{0.253} & 0.731 \\
    DeepSVDD            & 0.057 & 0.096 & 0.342 & 0.571 \\
    VLAConf             & 0.062 & \textbf{0.068} & 0.255 & \textbf{0.876} \\
    \bottomrule
    \end{tabular*}
    \vspace{-3mm}
\end{table}

\begin{table*}[!t]
    \centering
    \caption{Average robustness.}
    \vspace{-1mm}
    \label{tab:generalization_average}
    \footnotesize
    \renewcommand{\arraystretch}{0.92}
    \begin{tabular*}{\textwidth}{@{\extracolsep{\fill}}lllcccc}
    \toprule
    \textbf{Benchmark} & \textbf{Backbone} & \textbf{Method} & \textbf{ECE $\downarrow$} & \textbf{Brier $\downarrow$} & \textbf{NLL $\downarrow$} & \textbf{Success Rate} \\
    \midrule
    \multirow{3}{*}{LIBERO-Pro}
        & OpenVLA-OFT          & ConfidenceVLA & 0.0255 & 0.1645 & 0.4881 & 0.5055 \\
        & OpenVLA-OFT          & VLAConf          & 0.0303 & 0.0452 & 0.1512 & 0.5037 \\
        & \boldmath$\pi^{0.5}$ & VLAConf          & 0.0200 & 0.0250 & 0.3100 & 0.6722 \\
    \addlinespace
    \multirow{3}{*}{LIBERO-Plus}
        & OpenVLA-OFT          & ConfidenceVLA & 0.0355 & 0.1752 & 0.5197 & 0.3396 \\
        & OpenVLA-OFT          & VLAConf          & 0.0313 & 0.0325 & 0.1524 & 0.3405 \\
        & \boldmath$\pi^{0.5}$ & VLAConf          & 0.0433 & 0.0412 & 0.2824 & 0.5352 \\
    \bottomrule
    \end{tabular*}%
    \vspace{-4mm}
\end{table*}

\noindent\textbf{(1) VLAConf improves probability quality while remaining efficient.} On OpenVLA-OFT, where all baselines are applicable, VLAConf offers the best trade-off between calibration quality and inference cost. TokenProb is the cheapest baseline because it directly reuses the policy's own action-token probabilities, but its averaged Brier score and NLL are worse than those of VLAConf. PCA-kmeans is also consistently weaker, indicating that a simple clustering distance in frozen feature space is not sufficient. ConfidenceVLA is much more expensive because it repeatedly queries the VLA under a prompt ensemble, taking $712.9$\,ms per confidence query on average. In contrast, VLAConf adds only a lightweight head on top of frozen policy features and runs at $64.9$\,ms, essentially matching TokenProb ($65.1$\,ms) while being about $11\times$ faster than ConfidenceVLA. Despite this efficiency gap, VLAConf is strongest under the proper scoring rules in the averaged OpenVLA-OFT comparison. ConfidenceVLA still obtains the lowest online ECE, but this does not imply a more discriminative confidence signal: its prompt-ensemble score is concentrated and then Platt-scaled close to the empirical success rate, which can reduce bin-level calibration error while leaving little separation between successful and failed rollouts. The joint improvement on Brier score and NLL therefore provides stronger evidence that VLAConf produces sharper per-rollout probabilities rather than only matching the average success rate.

\noindent\textbf{(2) VLAConf provides more informative task-success confidence.} Table~\ref{tab:additional_confidence_baselines} reports the macro-averaged comparison with simple and representation-based predictors on $\pi^{0.5}$. VLAConf obtains the lowest Brier score and the highest failure AUROC while remaining competitive in NLL, indicating that its confidence estimates combine accurate outcome probabilities with stronger separation between successful and failed rollouts. Although Success Prior obtains a lower ECE, its chance-level failure AUROC shows that matching the average success frequency does not identify individual failures. Compared with Logistic Regression and DeepSVDD, VLAConf therefore provides the strongest overall balance between probability quality and failure discrimination.

\noindent\textbf{(3) VLAConf supports flow-matching policies.} VLAConf runs unchanged on both the discrete-action OpenVLA-OFT backbone and the continuous-action $\pi^{0.5}$ backbone, while TokenProb and ConfidenceVLA are unavailable for $\pi^{0.5}$ because they require discrete action-token probabilities. On $\pi^{0.5}$, VLAConf remains applicable on the LIBERO average, with pre-execution NLL of $0.3141$ and online NLL of $0.2501$. Compared with VLAConf-NoStep, it also gives lower Brier score and NLL in both regimes, showing that action-agnostic applicability alone is not enough; the representation-level score still benefits from phase conditioning. More broadly, because VLAConf only assumes access to frozen internal representations rather than a particular action parameterization, the same design can naturally extend to recent World Action Model policies that expose visual or dynamics features for control and planning~\cite{kim2026cosmos,li2026causal}.

\noindent\textbf{(4) Online execution evidence is often helpful, but its benefit is not uniform.} Comparing the pre-execution and online columns in Table~\ref{tab:confvla_alignment}, VLAConf benefits from observing the rollout prefix on OpenVLA-OFT: at the fixed-$50\%$ checkpoint, averaged Brier score and NLL both improve. The VLAConf-NoStep ablation further clarifies where this gain comes from. It keeps the same frozen features and success-support score-learning objective, yet removing the step encoder weakens Brier score and NLL in most settings, especially in online OpenVLA-OFT results. This suggests that execution evidence must be interpreted relative to task progress rather than treated as an unordered set of familiar or unfamiliar states. Figure~\ref{fig:online_temporal_calibration} shows a more nuanced temporal picture: some backbone--aggregator combinations improve substantially once task-relevant progress becomes visible, whereas others remain flat or degrade near the terminal stage. Thus, online evidence can sharpen confidence estimates, but the magnitude and timing of the benefit depend on both the backbone and the aggregation rule.

Together, these observations show that frozen VLA representations provide a strong basis for task-success confidence. VLAConf combines accurate probability estimates with strong failure discrimination, remains applicable to both discrete- and continuous-action policies, and can become more informative as online execution evidence accumulates.

% Declare these floats before the next subsection so IEEE's top-float queue
% places the selective-assistance figure and table together on page 7.
\begin{figure}[!t]
    \centering
    \includegraphics[width=0.98\columnwidth]{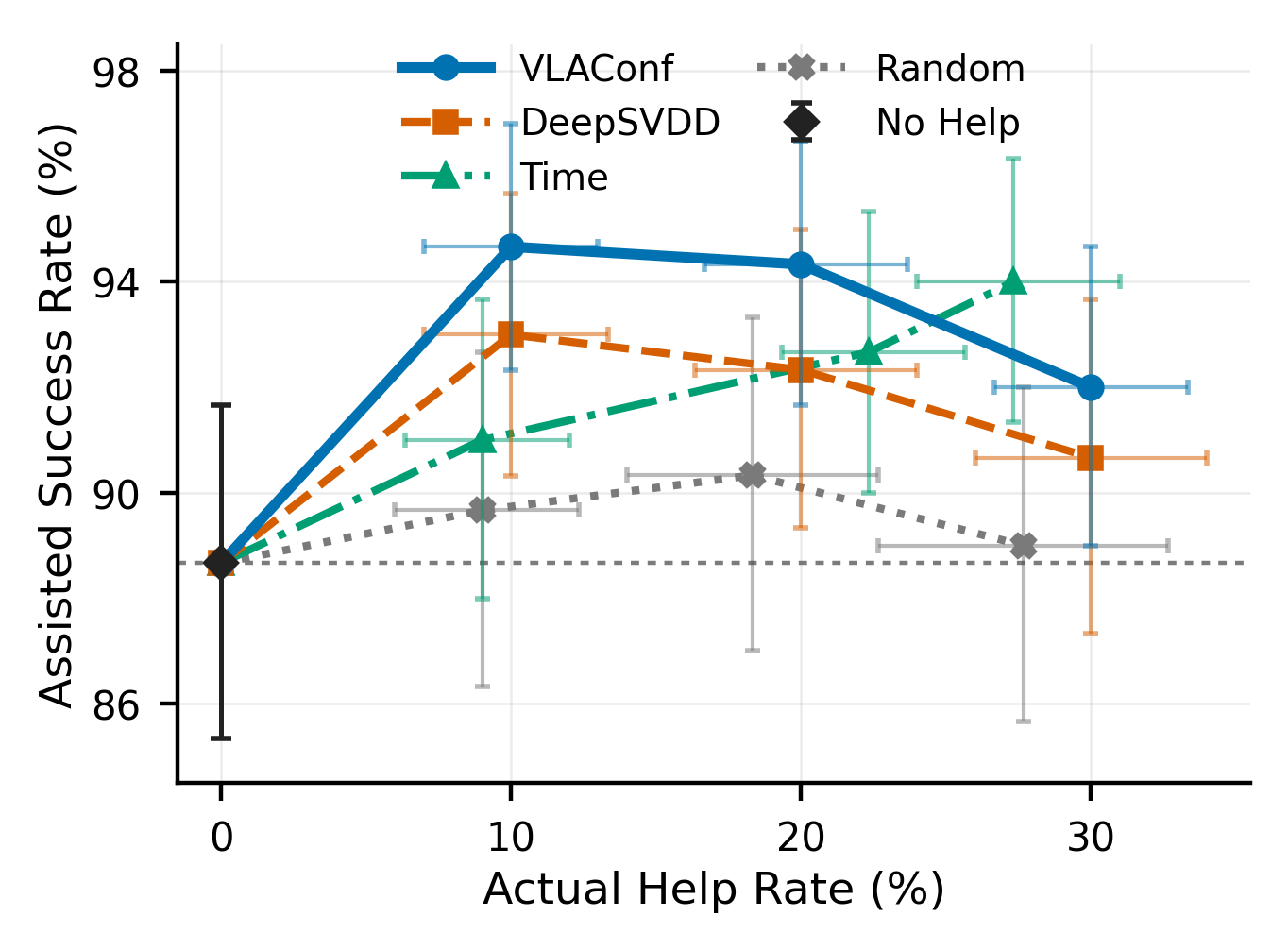}
    \caption{Selective assistance on LIBERO-Goal. Curves show assisted success rate (ASR) against actual help rate (HR); the shared No Help point is shown as a reference. Error bars are task-stratified bootstrap $95\%$ confidence intervals for both axes ($N{=}300$ per point).}
    \label{fig:assistance_budget_curve}
    \vspace{-4mm}
\end{figure}
% Source: logs/revision_baselines/assistance_analysis/assistance_curve_points.csv
% and logs/revision_baselines/assistance_analysis/assistance_results.json

\begin{table}[!t]
    \centering
    \caption{Selective assistance results on LIBERO-Goal.}
    \label{tab:assistance_summary}
    \footnotesize
    \setlength{\tabcolsep}{3.0pt}
    \renewcommand{\arraystretch}{1.03}
    \begin{tabular*}{\columnwidth}{@{\extracolsep{\fill}}lccc@{}}
    \toprule
    \textbf{Method} & \textbf{Actual HR (\%)} & \textbf{ASR (\%)} & \textbf{pAUC$_{0{:}30}$ $\uparrow$} \\
    \midrule
    VLAConf  & 10.0 & \textbf{94.7} & \textbf{0.931} \\
    DeepSVDD & 10.0 & 93.0 & 0.917 \\
    Time     & 9.0  & 91.0 & 0.917 \\
    Random   & 9.0  & 89.7 & 0.896 \\
    \bottomrule
    \end{tabular*}
    \vspace{-3mm}
\end{table}

\subsection{Robustness under Shifted LIBERO Suites}\label{exp:generalization}

The standard LIBERO results establish the behavior of VLAConf under the original benchmark distribution. We next evaluate on LIBERO-Pro and LIBERO-Plus to test whether the confidence signal remains useful under systematic distribution shifts, including task, object, layout, camera, robot-state, lighting, background, and sensor perturbations. Following the evaluation protocol above, each shifted suite uses lightweight per-suite Platt calibration before held-out evaluation. To keep the main text focused, Table~\ref{tab:generalization_average} reports average online execution results across perturbation families; the full perturbation-level pre-execution and online results are provided in Appendix Tables~\ref{tab:libero_pro_full} and~\ref{tab:libero_plus_full}.

Table~\ref{tab:generalization_average} shows that the main advantage of VLAConf persists under distribution shift after lightweight target-suite calibration. On OpenVLA-OFT, VLAConf and ConfidenceVLA have nearly identical success rates on both shifted suites, yet VLAConf substantially improves the proper scoring rules on both LIBERO-Pro and LIBERO-Plus. The ECE comparison is mixed---slightly worse on LIBERO-Pro but better on LIBERO-Plus---consistent with the standard-LIBERO results, where lower ECE alone does not necessarily indicate sharper per-rollout probabilities. The same representation-level head also remains applicable to the continuous-action $\pi^{0.5}$ backbone, where action-token-based confidence methods are unavailable.

The full results in Appendix Tables~\ref{tab:libero_pro_full} and~\ref{tab:libero_plus_full} further show that the benefit of online execution evidence largely survives distribution shift. For OpenVLA-OFT, moving from pre-execution to online execution reduces both Brier score and NLL on LIBERO-Pro and LIBERO-Plus. The same pattern appears in Brier score for $\pi^{0.5}$, although its LIBERO-Pro NLL slightly increases. The perturbation-level breakdown also exposes residual failure modes rather than hiding them in the average: for $\pi^{0.5}$, semantic and lighting shifts produce high online NLL despite high success rates, suggesting that a small number of overconfident tail failures can still dominate likelihood under certain shifts. Overall, these shifted-suite results indicate that VLAConf remains informative beyond the original benchmark distribution while preserving the main benefits observed in the standard setting.

\subsection{Selective Expert Assistance}\label{exp:assistance}

We next examine whether task-success confidence can support a downstream decision. On LIBERO-Goal, a low-confidence trigger requests a single handoff to a frozen expert for up to four action chunks, after which control returns to the deployed $\pi^{0.5}$ policy. We use a fixed set of $300$ test rollouts, with $30$ rollouts for each of the ten tasks, and terminate an episode once success is reached. For the compact main-text comparison, we show VLAConf together with DeepSVDD, a fixed-time trigger, and a random episode-level trigger.

To expose the assistance trade-off over a controlled range of budgets, the confidence-based methods rank complete passive test trajectories by their maximum failure-risk score and select exact empirical coverages near $10\%$, $20\%$, and $30\%$. This construction does not use test outcomes, but it does use each complete trajectory to determine episode-level coverage; we therefore treat it as a retrospective empirical-coverage analysis rather than a leak-free online deployment protocol. Time uses a fixed chunk index, while Random uses nested subsets induced by a fixed episode-level random score. All curves are plotted against the realized, rather than nominal, help rate.

Table~\ref{tab:assistance_summary} reports the low-budget operating point, corresponding to a nominal $10\%$ budget, together with the normalized pAUC over actual HR from $0$ to $30\%$. When the highest realized point falls slightly below $30\%$, we interpolate to the shared Always Help reference before computing pAUC. VLAConf reaches $94.7\%$ ASR at $10.0\%$ actual HR, compared with $88.7\%$ without assistance, and obtains the highest pAUC among the four methods. DeepSVDD is the strongest alternative at the low-budget point, while DeepSVDD and Time obtain the same pAUC. Figure~\ref{fig:assistance_budget_curve} also shows that the curves are non-monotonic: Time reaches $94.0\%$ ASR at $27.3\%$ actual HR, whereas VLAConf reaches $92.0\%$ at $30.0\%$ HR. The result therefore provides evidence of downstream utility on LIBERO-Goal under the evaluated protocol, rather than uniform superiority at every operating point or across suites. Complete curve values, confidence intervals, and additional routing baselines are reported in Appendix~\ref{app:assistance_results}.

\subsection{Real-World Robot Experiments}\label{exp:realworld}

\begin{figure}[!t]
    \centering
    \includegraphics[width=\columnwidth,trim=15 15 15 15,clip]{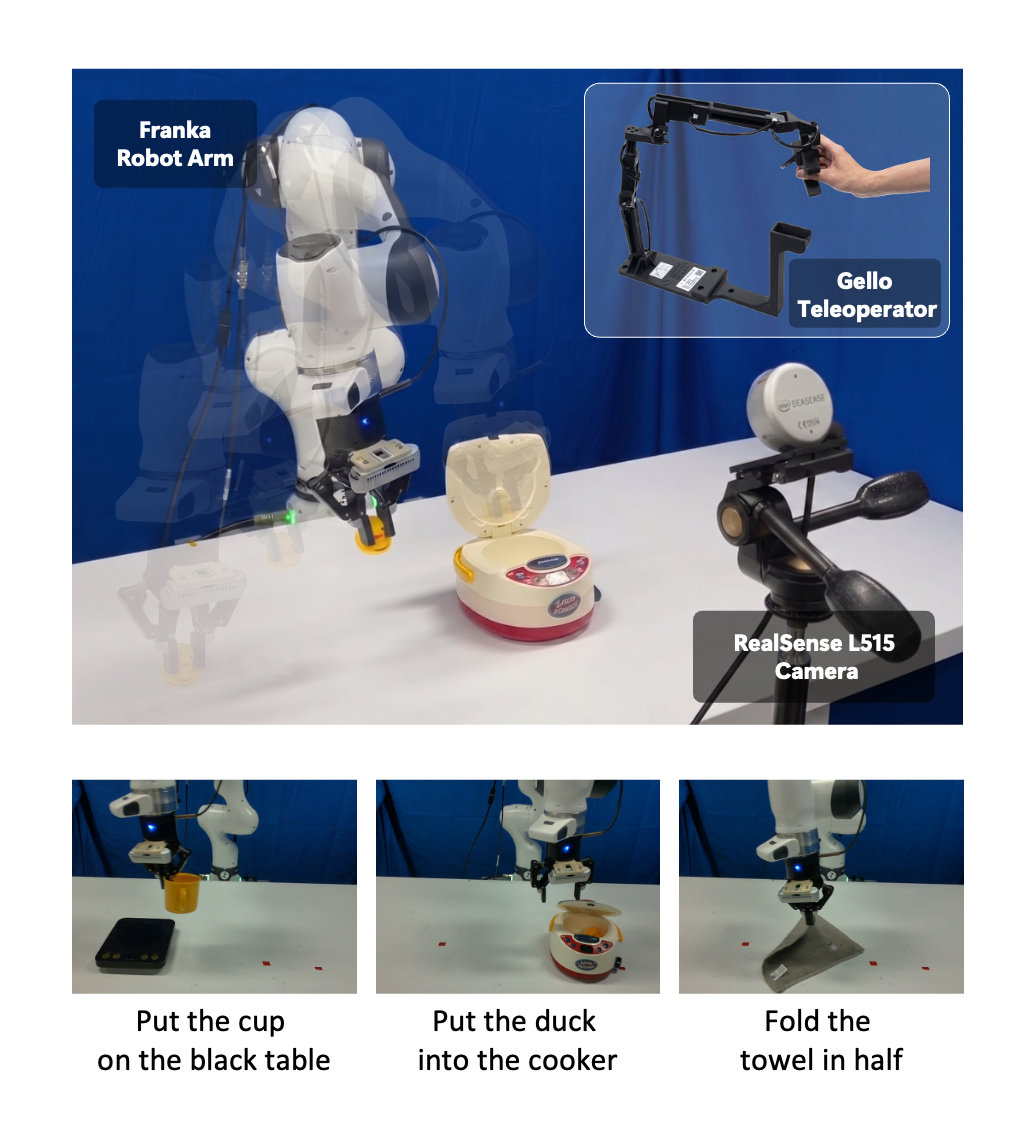}
    \vspace{-5mm}
    \caption{Real-world experimental setup and evaluation tasks. Tasks: cup-on-table, duck-in-cooker, and towel-folding.}
    \vspace{-4mm}
    \label{fig:real_robot_setup}
\end{figure}

\begin{table}[!t]
    \centering
    \caption{Real-world results.}
    \label{tab:real_robot_results}
    \footnotesize
    \setlength{\tabcolsep}{3.0pt}
    \renewcommand{\arraystretch}{1.05}
    \begin{tabular*}{\columnwidth}{@{\extracolsep{\fill}}lcccc@{}}
    \toprule
    \textbf{Task} & \textbf{Success} & \textbf{ECE $\downarrow$} & \textbf{Brier $\downarrow$} & \textbf{NLL $\downarrow$} \\
    \midrule
    Cup on Black Table  & 48/50 & 0.1011 & 0.0527 & 0.2361 \\
    Duck into Cooker    & 43/53 & 0.0528 & 0.1545 & 0.4888 \\
    Fold Towel          & 24/48 & 0.1584 & 0.2713 & 0.7406 \\
    \midrule
    Average             & 115/151 & 0.1041 & 0.1595 & 0.4885 \\
    \bottomrule
    \end{tabular*}
    \vspace{-6mm}
\end{table}

Finally, we evaluate whether the representation-level confidence signal remains useful beyond simulation by deploying VLAConf on a physical robot platform. As shown in Figure~\ref{fig:real_robot_setup}, the setup uses a Franka robot arm with a RealSense L515 camera, and demonstrations are collected through a Gello teleoperator. We consider three language-conditioned manipulation tasks: cup-on-table, duck-in-cooker, and towel-folding. These tasks cover rigid, articulated, and deformable manipulation, providing a deployment-facing complement to the benchmark studies above.

Table~\ref{tab:real_robot_results} summarizes task-success confidence over 151 physical rollouts. VLAConf remains effective in the real-world setting, with average ECE, Brier score, and NLL of $0.1041$, $0.1595$, and $0.4885$ across the three tasks. The cup-on-table task reaches $48/50$ successes and obtains the strongest proper-scoring-rule results, with Brier score $0.0527$ and NLL $0.2361$. Duck-in-cooker remains well calibrated in terms of ECE ($0.0528$), but exhibits higher Brier score and NLL. Towel-folding is the most challenging task in this set, achieving $24/48$ successes and the largest error on all three calibration metrics. These results show that the confidence signal remains informative on physical rollouts while also reflecting the increased uncertainty of harder real-world manipulation tasks. The corresponding real-world calibration trends across task completion are provided in Appendix Figure~\ref{fig:realworld_temporal_calibration}.

\section{Conclusion} \label{conclusion}
\vspace*{-0.01in}
We propose VLAConf, a representation-level confidence framework for Vision-Language-Action models that separates successful-demonstration support scoring from outcome-supervised probability calibration. Compared with existing confidence methods based on action-token probabilities or repeated sampling, VLAConf derives calibrated task-success confidence from frozen internal features and applies naturally to both discrete- and continuous-action backbones. Experiments on standard LIBERO suites and shifted LIBERO-Pro and LIBERO-Plus benchmarks demonstrate that VLAConf provides more informative task-success probabilities while remaining efficient, and that online evidence can further improve confidence quality during execution. Real-robot experiments further validate VLAConf.

\balance
\bibliographystyle{IEEEtran}
\bibliography{root}

\FloatBarrier
\clearpage
\onecolumn
\appendices
\section{}

\subsection{Implementation Details}\label{app:implementation_details}

The CFN head is a two-layer MLP applied on top of frozen VLA features, with masked-mean pooling over visual and language tokens and a small MLP encoder for proprioceptive state. The step descriptor consists of a learned embedding $\mathrm{Emb}(\hat{k})$ together with a normalized progress scalar $\hat{k}/(K{-}1)$, where $\hat{k}=\mathrm{clip}(k{-}1,0,K{-}1)$ and $K$ is a fixed hyperparameter ($K{=}256$ for both $\pi^{0.5}$ and OpenVLA-OFT). The descriptor modulates the CFN success-support head through a FiLM affine transform with a post-affine LayerNorm and a sigmoid residual gate. Coin-vector dimension is $d=64$. We train the success-support head on successful demonstration features with AdamW for the same number of optimization steps across backbones and suites. Within VLAConf, only the per-suite Platt calibration consumes outcome labels from successful and failed rollouts. Unless otherwise stated, temporal analyses use the running mean of step-wise scores, while the main scalar online results use max aggregation as described below.

\noindent\textbf{Online evaluation.} For the $\pi^{0.5}$ comparison, each suite contains $500$ rollouts, with $10\%$ used for outcome-stratified calibration and the remaining $450$ for evaluation. We report max-aggregated confidence at the $50\%$ execution checkpoint. Following the original protocol, progress on successful rollouts is normalized by the first-success time; these fixed-progress results are therefore retrospective rather than strictly causal deployment estimates. All methods use suite-matched splits, although their deterministic replays are not paired episode by episode.

\subsection{Additional Analyses}

This section presents additional analyses that support the main experimental findings while keeping the main text focused. We first examine whether the raw inverse success-support score reflects execution difficulty before calibration. Following that, we report temporal calibration trends for the real-robot study, providing a more detailed view of how the confidence estimates evolve as task execution proceeds.

\subsubsection{Prefix-Score Progress Analysis}

\begin{figure}[H]
    \centering
    \includegraphics[width=0.74\textwidth]{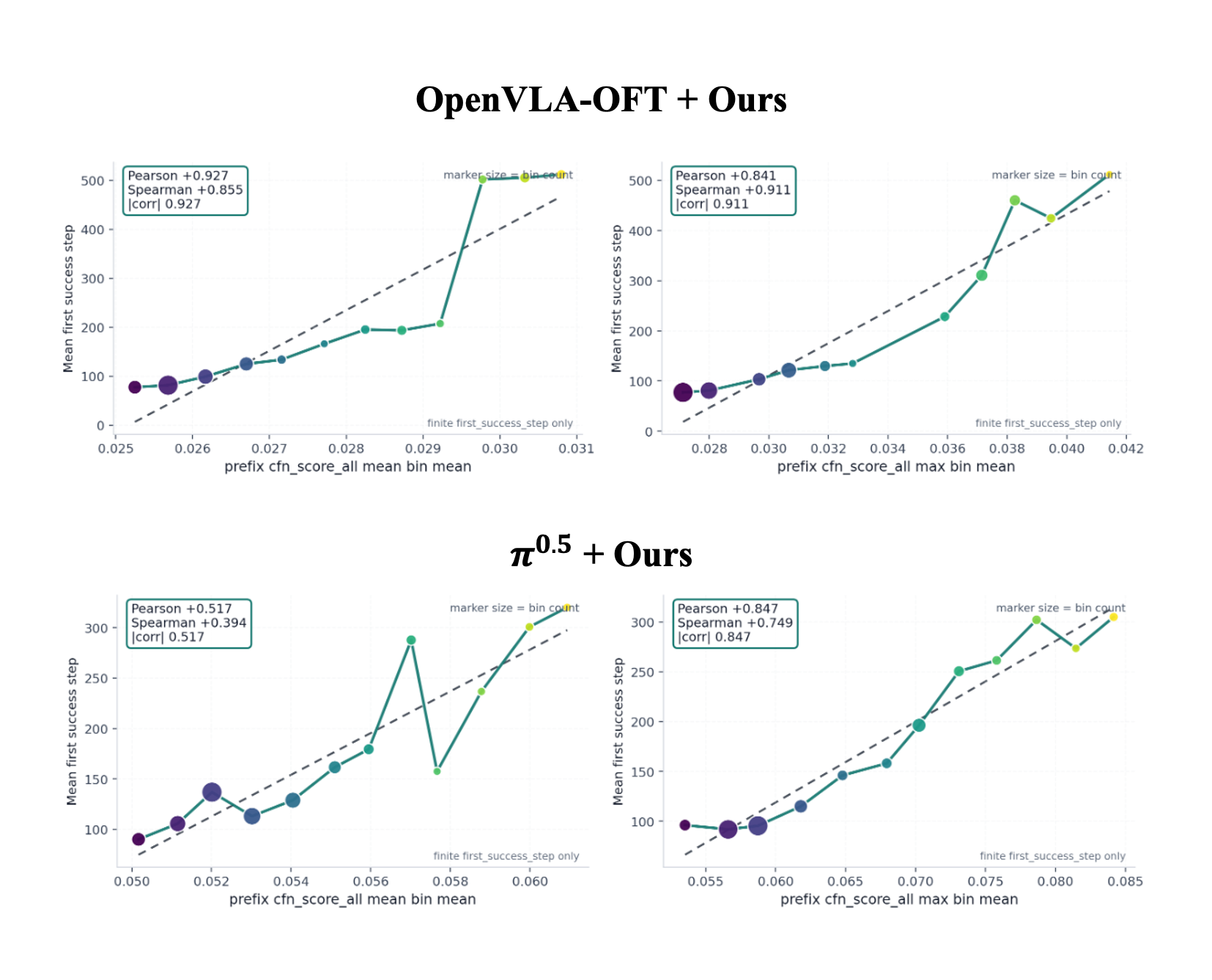}
    \caption{Prefix-score progress analysis. Successful rollouts are bucketed by their prefix CFN score, using mean and max aggregation, and each marker reports the mean first-success step in that bucket; marker size denotes the bucket count. Higher CFN scores correlate with later first-success steps, suggesting that the uncalibrated support-deviation signal tracks execution difficulty.}
    \label{fig:cfn_success_step}
\end{figure}

Figure~\ref{fig:cfn_success_step} analyzes the uncalibrated CFN inverse success-support signal on successful rollouts before Platt scaling. Successful trajectories with larger prefix scores tend to reach their first successful completion later under both mean and max aggregation. This pattern suggests that the score captures more than a binary support-membership signal: even among eventual successes, it reflects the apparent difficulty or delay of the execution process.

\subsubsection{Real-World Calibration Across Task Completion}

\begin{figure}[H]
    \centering
    \includegraphics[width=\textwidth]{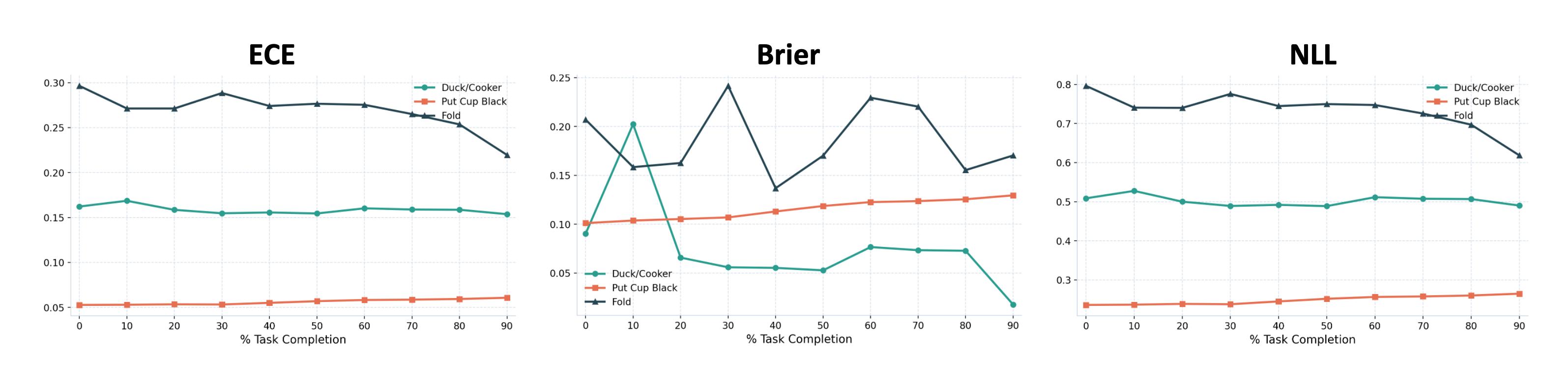}
    \caption{Real-world calibration trends across task completion for the three evaluation tasks. ECE, Brier score, and NLL are reported at increasing prefix lengths for cup-on-table, duck-in-cooker, and towel-folding.}
    \label{fig:realworld_temporal_calibration}
\end{figure}

Figure~\ref{fig:realworld_temporal_calibration} complements the scalar real-world results in Table~\ref{tab:real_robot_results} by showing how calibration evolves during execution. The temporal profiles differ across tasks: the easiest task remains comparatively stable, whereas the harder manipulation tasks exhibit larger residual errors. This trend is consistent with the task-wise summary reported in the main paper.

\subsection{Full Standard LIBERO Results}

This section provides the suite-level breakdown underlying the averaged standard-LIBERO comparison in Table~\ref{tab:confvla_alignment}. As shown in Table~\ref{tab:libero_standard_full}, the full results report both pre-execution and online-execution calibration metrics for each backbone and benchmark suite, together with the policy success rate and average inference time.

\begin{table}[H]
    \centering
    \caption{Full standard LIBERO comparison under pre-execution and online execution.}
    \label{tab:libero_standard_full}
    \footnotesize
    \setlength{\tabcolsep}{2.1pt}
    \renewcommand{\arraystretch}{1.02}
    \begin{tabular*}{\textwidth}{@{\extracolsep{\fill}}lllcccccccc@{}}
    \toprule
    \multirow{2}{*}{\textbf{Backbone}} &
    \multirow{2}{*}{\textbf{Suite}} &
    \multirow{2}{*}{\textbf{Method}} &
    \multicolumn{3}{c}{\textbf{Pre-Execution}} &
    \multicolumn{3}{c}{\textbf{Online Execution}} &
    \multirow{2}{*}{\textbf{Succ. (\%)}} &
    \multirow{2}{*}{\textbf{Avg. Inf. Time (ms)}} \\
    \cmidrule(lr){4-6} \cmidrule(lr){7-9}
    & & & \textbf{ECE $\downarrow$} & \textbf{Brier $\downarrow$} & \textbf{NLL $\downarrow$} & \textbf{ECE $\downarrow$} & \textbf{Brier $\downarrow$} & \textbf{NLL $\downarrow$} & & \\
    \midrule
    \multirow{15}{*}{OpenVLA-OFT}
    & \multirow{5}{*}{LIBERO-Spatial}
    & PCA-kmeans    & 0.2819 & 0.3373 & 1.2805 & 0.4779 & 0.4650 & 1.3961 & 73.0 & 62.5 \\
    &   & TokenProb     & 0.2112 & 0.2315 & 0.6541 & 0.0279 & 0.1758 & 0.5327 & 76.0 & 65.4 \\
    &   & ConfidenceVLA & \textbf{0.0182} & \textbf{0.1820} & \textbf{0.5498} & \textbf{0.0215} & 0.1804 & 0.5448 & 76.0 & 710.5 \\
    &   & VLAConf-NoStep      & 0.0406 & 0.1925 & 0.5743 & 0.0730 & 0.1907 & 0.5685 & 74.0 & 65.1 \\
    &   & VLAConf          & 0.0330 & 0.1880 & 0.5649 & 0.2128 & \textbf{0.1196} & \textbf{0.3789} & 74.6 & 64.9 \\
    \addlinespace
    & \multirow{5}{*}{LIBERO-Object}
    & PCA-kmeans    & 0.2769 & 0.3083 & 1.0499 & 0.2846 & 0.3122 & 1.0341 & 69.2 & 63.9 \\
    &   & TokenProb     & 0.1344 & 0.2402 & 0.6735 & 0.0962 & 0.2239 & 0.6529 & 68.8 & 64.4 \\
    &   & ConfidenceVLA & \textbf{0.0232} & 0.2136 & 0.6183 & \textbf{0.0428} & 0.2113 & 0.6130 & 68.8 & 711.7 \\
    &   & VLAConf-NoStep      & 0.0615 & 0.2181 & 0.6273 & 0.0878 & 0.2096 & 0.6084 & 67.6 & 63.3 \\
    &   & VLAConf          & 0.0351 & \textbf{0.2071} & \textbf{0.6035} & 0.1102 & \textbf{0.1950} & \textbf{0.5730} & 69.8 & 64.9 \\
    \addlinespace
    & \multirow{5}{*}{LIBERO-Goal}
    & PCA-kmeans    & 0.3506 & 0.2853 & 0.9624 & 0.3442 & 0.2759 & 0.9205 & 88.4 & 64.8 \\
    &   & TokenProb     & 0.2912 & 0.1944 & 0.5779 & 0.1525 & 0.1296 & 0.4356 & 88.2 & 65.4 \\
    &   & ConfidenceVLA & 0.0674 & 0.1150 & 0.4203 & \textbf{0.0186} & 0.1023 & 0.3545 & 88.2 & 716.5 \\
    &   & VLAConf-NoStep      & \textbf{0.0208} & 0.0967 & 0.3428 & 0.0375 & 0.0966 & 0.3420 & 89.0 & 65.1 \\
    &   & VLAConf          & 0.0339 & \textbf{0.0892} & \textbf{0.3289} & 0.0335 & \textbf{0.0073} & \textbf{0.0485} & 90.2 & 64.9 \\
    \midrule
    \multirow{12}{*}{\boldmath$\pi^{0.5}$}
    & \multirow{4}{*}{LIBERO-Spatial}
    & PCA-kmeans    & 0.3623 & 0.2498 & 0.7316 & 0.3835 & 0.2814 & 0.8247 & 90.2 & 163.0 \\
    &   & ConfidenceVLA & / & / & / & / & / & / & / & / \\
    &   & VLAConf-NoStep      & 0.0634 & 0.0865 & 0.3399 & 0.0443 & 0.0838 & 0.3107 & 90.8 & 164.6 \\
    &   & VLAConf          & \textbf{0.0352} & \textbf{0.0723} & \textbf{0.2768} & \textbf{0.0265} & \textbf{0.0713} & \textbf{0.2703} & 91.8 & 164.7 \\
    \addlinespace
    & \multirow{4}{*}{LIBERO-Object}
    & PCA-kmeans    & 0.4703 & 0.4378 & 1.4658 & 0.4388 & 0.4172 & 1.3327 & 92.6 & 164.3 \\
    &   & ConfidenceVLA & / & / & / & / & / & / & / & / \\
    &   & VLAConf-NoStep      & \textbf{0.0619} & 0.0909 & 0.3456 & 0.0552 & 0.0888 & 0.3327 & 90.8 & 163.7 \\
    &   & VLAConf          & 0.0735 & \textbf{0.0835} & \textbf{0.3375} & \textbf{0.0259} & \textbf{0.0759} & \textbf{0.2902} & 92.2 & 165.2 \\
    \addlinespace
    & \multirow{4}{*}{LIBERO-Goal}
    & PCA-kmeans    & 0.4178 & 0.3407 & 1.1173 & 0.3936 & 0.3197 & 1.0315 & 91.2 & 163.3 \\
    &   & ConfidenceVLA & / & / & / & / & / & / & / & / \\
    &   & VLAConf-NoStep      & 0.0484 & 0.0919 & 0.3307 & \textbf{0.0348} & 0.0926 & 0.3348 & 89.8 & 165.0 \\
    &   & VLAConf          & \textbf{0.0024} & \textbf{0.0906} & \textbf{0.3281} & 0.1020 & \textbf{0.0531} & \textbf{0.1898} & 90.0 & 164.1 \\
    \bottomrule
    \end{tabular*}
\end{table}

\clearpage

\subsection{Full Perturbation-Level Robustness Results}

This section further details the perturbation-level robustness results summarized in Table~\ref{tab:generalization_average}. Tables~\ref{tab:libero_pro_full} and~\ref{tab:libero_plus_full} report both pre-execution and online-execution calibration for each perturbation family, showing where online evidence provides the strongest gains and where residual failure modes remain visible after averaging.

\begin{table}[H]
    \centering
    \caption{Full LIBERO-Pro robustness results under pre-execution and online execution.}
    \label{tab:libero_pro_full}
    \footnotesize
    \setlength{\tabcolsep}{2.0pt}
    \renewcommand{\arraystretch}{1.02}
    \begin{tabular*}{\textwidth}{@{\extracolsep{\fill}}ll l ccc ccc c@{}}
        \toprule
        \multirow{2}{*}{\textbf{Perturbation}} & \multirow{2}{*}{\textbf{Backbone}} & \multirow{2}{*}{\textbf{Method}} &
        \multicolumn{3}{c}{\textbf{Pre-Execution}} & \multicolumn{3}{c}{\textbf{Online Execution}} &
        \multirow{2}{*}{\textbf{Success Rate}} \\
        \cmidrule(lr){4-6} \cmidrule(lr){7-9}
        & & & \textbf{ECE $\downarrow$} & \textbf{Brier $\downarrow$} & \textbf{NLL $\downarrow$}
          & \textbf{ECE $\downarrow$} & \textbf{Brier $\downarrow$} & \textbf{NLL $\downarrow$} & \\
        \midrule
        \multirow{3}{*}{Object}
         & OpenVLA-OFT          & ConfidenceVLA & 0.0367 & 0.2265 & 0.6457 & 0.0034 & 0.2245 & 0.6413 & 0.6593 \\
         & OpenVLA-OFT          & VLAConf          & 0.0354 & 0.2258 & 0.6436 & 0.0157 & 0.0503 & 0.1740 & 0.6467 \\
         & \boldmath$\pi^{0.5}$ & VLAConf          & 0.0400 & 0.1211 & 0.4112 & 0.0303 & 0.0380 & 0.1203 & 0.8626 \\
        \midrule
        \multirow{3}{*}{Position}
         & OpenVLA-OFT          & ConfidenceVLA & 0.0058 & 0.2456 & 0.6844 & 0.0603 & 0.2327 & 0.6586 & 0.5680 \\
         & OpenVLA-OFT          & VLAConf          & 0.0401 & 0.2405 & 0.6739 & 0.0320 & 0.0513 & 0.1490 & 0.5647 \\
         & \boldmath$\pi^{0.5}$ & VLAConf          & 0.0370 & 0.1454 & 0.4670 & 0.0118 & 0.0242 & 0.0758 & 0.8247 \\
        \midrule
        \multirow{3}{*}{Semantic}
         & OpenVLA-OFT          & ConfidenceVLA & 0.0271 & 0.1761 & 0.5370 & 0.0362 & 0.1790 & 0.5466 & 0.7727 \\
         & OpenVLA-OFT          & VLAConf          & 0.0362 & 0.1741 & 0.5326 & 0.0705 & 0.0756 & 0.2560 & 0.7793 \\
         & \boldmath$\pi^{0.5}$ & VLAConf          & 0.0031 & 0.0332 & 0.1522 & 0.0341 & 0.0341 & 0.9415 & 0.9660 \\
        \midrule
        \multirow{3}{*}{Task}
         & OpenVLA-OFT          & ConfidenceVLA & 0.0176 & 0.0234 & 0.1238 & 0.0021 & 0.0217 & 0.1057 & 0.0220 \\
         & OpenVLA-OFT          & VLAConf          & 0.0031 & 0.0232 & 0.1125 & 0.0031 & 0.0037 & 0.0258 & 0.0240 \\
         & \boldmath$\pi^{0.5}$ & VLAConf          & 0.0022 & 0.0342 & 0.1515 & 0.0037 & 0.0037 & 0.1023 & 0.0353 \\
        \midrule\midrule
        \multirow{3}{*}{\textbf{Average}}
         & OpenVLA-OFT          & ConfidenceVLA & \textbf{0.0218} & \textbf{0.1679} & \textbf{0.4977} & \textbf{0.0255} & \textbf{0.1645} & \textbf{0.4881} & \textbf{0.5055} \\
         & OpenVLA-OFT          & VLAConf          & \textbf{0.0287} & \textbf{0.1659} & \textbf{0.4907} & \textbf{0.0303} & \textbf{0.0452} & \textbf{0.1512} & \textbf{0.5037} \\
         & \boldmath$\pi^{0.5}$ & VLAConf          & \textbf{0.0206} & \textbf{0.0835} & \textbf{0.2955} & \textbf{0.0200} & \textbf{0.0250} & \textbf{0.3100} & \textbf{0.6722} \\
        \bottomrule
    \end{tabular*}
\end{table}

\begin{table}[H]
    \centering
    \caption{Full LIBERO-Plus robustness results under pre-execution and online execution.}
    \label{tab:libero_plus_full}
    \footnotesize
    \setlength{\tabcolsep}{2.0pt}
    \renewcommand{\arraystretch}{1.02}
    \begin{tabular*}{\textwidth}{@{\extracolsep{\fill}}ll l ccc ccc c@{}}
        \toprule
        \multirow{2}{*}{\textbf{Perturbation}} & \multirow{2}{*}{\textbf{Backbone}} & \multirow{2}{*}{\textbf{Method}} &
        \multicolumn{3}{c}{\textbf{Pre-Execution}} & \multicolumn{3}{c}{\textbf{Online Execution}} &
        \multirow{2}{*}{\textbf{Success Rate}} \\
        \cmidrule(lr){4-6} \cmidrule(lr){7-9}
        & & & \textbf{ECE $\downarrow$} & \textbf{Brier $\downarrow$} & \textbf{NLL $\downarrow$}
          & \textbf{ECE $\downarrow$} & \textbf{Brier $\downarrow$} & \textbf{NLL $\downarrow$} & \\
        \midrule
        \multirow{3}{*}{Layout}
         & OpenVLA-OFT          & ConfidenceVLA & 0.0010 & 0.2498 & 0.6927 & 0.0381 & 0.2441 & 0.6813 & 0.4864 \\
         & OpenVLA-OFT          & VLAConf          & 0.0225 & 0.2415 & 0.6758 & 0.0161 & 0.0344 & 0.1401 & 0.4971 \\
         & \boldmath$\pi^{0.5}$ & VLAConf          & 0.0104 & 0.1629 & 0.5083 & 0.0270 & 0.0290 & 0.1087 & 0.7980 \\
        \midrule
        \multirow{3}{*}{Camera}
         & OpenVLA-OFT          & ConfidenceVLA & 0.0132 & 0.0346 & 0.1649 & 0.0010 & 0.0327 & 0.1468 & 0.0339 \\
         & OpenVLA-OFT          & VLAConf          & 0.0016 & 0.0311 & 0.1431 & 0.0075 & 0.0075 & 0.1699 & 0.0322 \\
         & \boldmath$\pi^{0.5}$ & VLAConf          & 0.0023 & 0.1370 & 0.4460 & 0.0521 & 0.0383 & 0.1658 & 0.1636 \\
        \midrule
        \multirow{3}{*}{Robot Init}
         & OpenVLA-OFT          & ConfidenceVLA & 0.0019 & 0.1049 & 0.3651 & 0.0056 & 0.1061 & 0.3706 & 0.1193 \\
         & OpenVLA-OFT          & VLAConf          & 0.0244 & 0.1085 & 0.3806 & 0.0197 & 0.0358 & 0.1674 & 0.1219 \\
         & \boldmath$\pi^{0.5}$ & VLAConf          & 0.0023 & 0.1292 & 0.4265 & 0.0656 & 0.0442 & 0.1925 & 0.1530 \\
        \midrule
        \multirow{3}{*}{Language}
         & OpenVLA-OFT          & ConfidenceVLA & 0.0021 & 0.2049 & 0.6000 & 0.0249 & 0.2059 & 0.6020 & 0.7114 \\
         & OpenVLA-OFT          & VLAConf          & 0.0023 & 0.2035 & 0.5971 & 0.0268 & 0.0310 & 0.1002 & 0.7149 \\
         & \boldmath$\pi^{0.5}$ & VLAConf          & 0.0877 & 0.2434 & 0.6821 & 0.0291 & 0.0384 & 0.1388 & 0.6334 \\
        \midrule
        \multirow{3}{*}{Light}
         & OpenVLA-OFT          & ConfidenceVLA & 0.0058 & 0.2268 & 0.6458 & 0.0543 & 0.2353 & 0.6648 & 0.3537 \\
         & OpenVLA-OFT          & VLAConf          & 0.0766 & 0.2320 & 0.6577 & 0.0286 & 0.0352 & 0.1532 & 0.3456 \\
         & \boldmath$\pi^{0.5}$ & VLAConf          & 0.0014 & 0.0356 & 0.1564 & 0.0371 & 0.0371 & 1.0260 & 0.9631 \\
        \midrule
        \multirow{3}{*}{Background}
         & OpenVLA-OFT          & ConfidenceVLA & 0.0201 & 0.2490 & 0.6913 & 0.0990 & 0.2593 & 0.7121 & 0.5006 \\
         & OpenVLA-OFT          & VLAConf          & 0.0147 & 0.2505 & 0.6941 & 0.1059 & 0.0607 & 0.2165 & 0.5006 \\
         & \boldmath$\pi^{0.5}$ & VLAConf          & 0.0224 & 0.1419 & 0.4579 & 0.0526 & 0.0550 & 0.1612 & 0.8285 \\
        \midrule
        \multirow{3}{*}{Noise}
         & OpenVLA-OFT          & ConfidenceVLA & 0.0296 & 0.1434 & 0.4626 & 0.0254 & 0.1428 & 0.4606 & 0.1719 \\
         & OpenVLA-OFT          & VLAConf          & 0.0096 & 0.1417 & 0.4573 & 0.0146 & 0.0230 & 0.1197 & 0.1710 \\
         & \boldmath$\pi^{0.5}$ & VLAConf          & 0.0095 & 0.1636 & 0.5087 & 0.0399 & 0.0462 & 0.1840 & 0.2066 \\
        \midrule\midrule
        \multirow{3}{*}{\textbf{Average}}
         & OpenVLA-OFT          & ConfidenceVLA & \textbf{0.0105} & \textbf{0.1733} & \textbf{0.5175} & \textbf{0.0355} & \textbf{0.1752} & \textbf{0.5197} & \textbf{0.3396} \\
         & OpenVLA-OFT          & VLAConf          & \textbf{0.0217} & \textbf{0.1727} & \textbf{0.5151} & \textbf{0.0313} & \textbf{0.0325} & \textbf{0.1524} & \textbf{0.3405} \\
         & \boldmath$\pi^{0.5}$ & VLAConf          & \textbf{0.0194} & \textbf{0.1448} & \textbf{0.4551} & \textbf{0.0433} & \textbf{0.0412} & \textbf{0.2824} & \textbf{0.5352} \\
        \bottomrule
    \end{tabular*}
\end{table}

\clearpage

\subsection{Full Selective Assistance Results}\label{app:assistance_results}

Table~\ref{tab:assistance_full} reports all operating points underlying the selective-assistance analysis in Figure~\ref{fig:assistance_budget_curve}, including the confidence and routing baselines omitted from the compact main-text comparison. Each point uses the same fixed set of $300$ LIBERO-Goal episodes. Confidence intervals are computed with $10{,}000$ task-stratified bootstrap resamples. The normalized pAUC is computed over actual HR from $0$ to $30\%$; when a method's highest realized point falls below $30\%$, we linearly interpolate to the shared Always Help reference at that boundary.

\begin{table}[H]
    \centering
    \caption{Full selective assistance results on LIBERO-Goal.}
    \label{tab:assistance_full}
    \footnotesize
    \setlength{\tabcolsep}{3.5pt}
    \renewcommand{\arraystretch}{1.02}
    \begin{tabular*}{\textwidth}{@{\extracolsep{\fill}}lccccc c@{}}
        \toprule
        \textbf{Method} & \textbf{Budget} & \textbf{Actual HR (\%)} & \textbf{HR 95\% CI} & \textbf{ASR (\%)} & \textbf{ASR 95\% CI} & \textbf{pAUC$_{0{:}30}$} \\
        \midrule
        \multirow{3}{*}{VLAConf}
          & $10\%$ & 10.0 & $[7.0, 13.0]$ & 94.7 & $[92.3, 97.0]$ & \multirow{3}{*}{0.931} \\
          & $20\%$ & 20.0 & $[16.7, 23.7]$ & 94.3 & $[91.7, 96.7]$ & \\
          & $30\%$ & 30.0 & $[26.7, 33.3]$ & 92.0 & $[89.0, 94.7]$ & \\
        \addlinespace
        \multirow{3}{*}{Logistic Regression}
          & $10\%$ & 10.0 & $[7.3, 12.7]$ & 90.7 & $[87.3, 93.7]$ & \multirow{3}{*}{0.904} \\
          & $20\%$ & 20.0 & $[17.0, 23.3]$ & 90.3 & $[87.3, 93.3]$ & \\
          & $30\%$ & 30.0 & $[26.0, 34.0]$ & 92.0 & $[89.3, 94.7]$ & \\
        \addlinespace
        \multirow{3}{*}{Mahalanobis}
          & $10\%$ & 10.0 & $[7.0, 13.0]$ & 90.3 & $[87.0, 93.3]$ & \multirow{3}{*}{0.901} \\
          & $20\%$ & 20.0 & $[16.3, 23.7]$ & 90.3 & $[87.3, 93.0]$ & \\
          & $30\%$ & 30.0 & $[26.3, 33.7]$ & 90.7 & $[87.7, 93.3]$ & \\
        \addlinespace
        \multirow{3}{*}{DeepSVDD}
          & $10\%$ & 10.0 & $[7.0, 13.3]$ & 93.0 & $[90.3, 95.7]$ & \multirow{3}{*}{0.917} \\
          & $20\%$ & 20.0 & $[16.3, 24.0]$ & 92.3 & $[89.3, 95.0]$ & \\
          & $30\%$ & 30.0 & $[26.0, 34.0]$ & 90.7 & $[87.3, 93.7]$ & \\
        \addlinespace
        \multirow{3}{*}{Energy}
          & $10\%$ & 10.0 & $[7.3, 12.7]$ & 89.3 & $[86.0, 92.3]$ & \multirow{3}{*}{0.901} \\
          & $20\%$ & 20.0 & $[16.7, 23.3]$ & 91.7 & $[88.7, 94.3]$ & \\
          & $30\%$ & 30.0 & $[26.3, 33.7]$ & 90.0 & $[87.0, 93.0]$ & \\
        \addlinespace
        \multirow{3}{*}{Random}
          & $10\%$ & 9.0 & $[6.0, 12.3]$ & 89.7 & $[86.3, 92.7]$ & \multirow{3}{*}{0.896} \\
          & $20\%$ & 18.3 & $[14.0, 22.7]$ & 90.3 & $[87.0, 93.3]$ & \\
          & $30\%$ & 27.7 & $[22.7, 32.7]$ & 89.0 & $[85.7, 92.0]$ & \\
        \addlinespace
        \multirow{3}{*}{Time}
          & $10\%$ & 9.0 & $[6.3, 12.0]$ & 91.0 & $[88.0, 93.7]$ & \multirow{3}{*}{0.917} \\
          & $20\%$ & 22.3 & $[19.3, 25.7]$ & 92.7 & $[90.0, 95.3]$ & \\
          & $30\%$ & 27.3 & $[24.0, 31.0]$ & 94.0 & $[91.3, 96.3]$ & \\
        \addlinespace
        \multirow{3}{*}{Task-Prior}
          & $10\%$ & 10.0 & $[10.0, 10.0]$ & 89.7 & $[86.3, 92.7]$ & \multirow{3}{*}{0.901} \\
          & $20\%$ & 20.0 & $[20.0, 20.0]$ & 89.7 & $[86.3, 92.7]$ & \\
          & $30\%$ & 30.0 & $[30.0, 30.0]$ & 93.3 & $[90.7, 95.7]$ & \\
        \midrule
        No Help & Reference & 0.0 & $[0.0, 0.0]$ & 88.7 & $[85.3, 91.7]$ & -- \\
        Always Help & Reference & 100.0 & $[100.0, 100.0]$ & 92.3 & $[89.3, 95.0]$ & -- \\
        Expert-only & Reference & -- & -- & 97.7 & $[96.0, 99.0]$ & -- \\
        \bottomrule
    \end{tabular*}
\end{table}

\end{document}